\documentclass[11pt,draftcls,onecolumn]{IEEEtran}

\usepackage{amsmath,amssymb,amsthm,epsfig,dsfont,color,subfigure,empheq,enumerate}
\usepackage{algorithmic, algorithm}

\DeclareMathOperator{\sign}{sign}

\DeclareMathOperator{\pr}{Pr}

\newtheorem{lemma}{Lemma}
\newtheorem{corollary}{Corollary}
\newtheorem{theorem}{Theorem}
\newtheorem{definition}{Definition}
\theoremstyle{remark}\newtheorem{remark}{Remark}

\newcommand{\mathbft}[1]{\ensuremath{\tilde{\mathbf{#1}}}}

\begin{document}
\title{Sparse Volterra and Polynomial Regression\\
Models: Recoverability and Estimation}

\author{Vassilis Kekatos,~\IEEEmembership{Member,~IEEE,} and Georgios B. Giannakis*,~\IEEEmembership{Fellow,~IEEE}%
\thanks{Part of the results of this work was presented at \cite{KeAnGia09}. Work was supported by the Marie Curie International Outgoing Fellowship No.~234914 within the $7$-th European Community Framework Programme; by NSF grants CCF-0830480, 1016605, and ECCS-0824007, 1002180; and by the QNRF-NPRP award 09-341-2-128. The authors are with the ECE Dept., University of Minnesota, Minneapolis, MN 55455, USA, Emails:\{kekatos,georgios\}@umn.edu.}}

\markboth{IEEE TRANSACTIONS ON SIGNAL PROCESSING (REVISED)}{Kekatos
and Giannakis: Sparse Volterra and Polynomial Regression Models:
Recoverability and Estimation}

\maketitle

\begin{abstract}
Volterra and polynomial regression models play a major role in
nonlinear system identification and inference tasks. Exciting
applications ranging from neuroscience to genome-wide association
analysis build on these models with the additional requirement of
parsimony. This requirement has high interpretative value, but
unfortunately cannot be met by least-squares based or kernel
regression methods. To this end, compressed sampling (CS)
approaches, already successful in linear regression settings, can
offer a viable alternative. The viability of CS for sparse Volterra
and polynomial models is the core theme of this work. A common
sparse regression task is initially posed for the two models.
Building on (weighted) Lasso-based schemes, an adaptive RLS-type
algorithm is developed for sparse polynomial regressions. The
identifiability of polynomial models is critically challenged by
dimensionality. However, following the CS principle, when these
models are sparse, they could be recovered by far fewer
measurements. To quantify the sufficient number of measurements for
a given level of sparsity, restricted isometry properties (RIP) are
investigated in commonly met polynomial regression settings,
generalizing known results for their linear counterparts. The merits
of the novel (weighted) adaptive CS algorithms to sparse polynomial
modeling are verified through synthetic as well as real data tests
for genotype-phenotype analysis.
\end{abstract}




\newpage

\section{Introduction}
Nonlinear systems with memory appear frequently in science and
engineering. Pertinent application areas include physiological and
biological processes \cite{Mar10}, power amplifiers \cite{BeBi83},
loudspeakers \cite{ZeKe10}, speech, and image models, to name a few;
see e.g., \cite{MaSi00}. If the nonlinearity is sufficiently smooth,
the Volterra series offers a well-appreciated model of the output
expressed as a polynomial expansion of the input using Taylor's
theorem \cite{PaPo77}. The expansion coefficients of order $P>1$ are
$P$-dimensional sequences of memory $L$ generalizing the
one-dimensional impulse response sequence encountered with linear
systems. However, polynomial expansions of nonlinear mappings go
beyond filtering. Polynomial regression aims at approximating a
multivariate nonlinear function via a polynomial expansion
\cite{HaTiFr09}. Apart from its extensive use for optical character
recognition and other classification tasks \cite{ScSm02},
(generalized) polynomial regression has recently emerged as a
valuable tool for revealing genotype-phenotype relationships in
genome-wide association (GWA) studies \cite{Co09}, \cite{WuCh09},
\cite{Xu07}, \cite{NaNo10}.

Volterra and polynomial regression models are jointly investigated
here. Albeit nonlinear, their input-output (I/O) relationship is
linear with respect to the unknown parameters, and can thus be
estimated via linear least-squares (LS) \cite{MaSi00},
\cite{HaTiFr09}. The major bottleneck is the ``curse of
dimensionality,'' since the number of regression coefficients $M$
grows as $\mathcal{O}(L^P)$. This not only raises computational and
numerical stability challenges, but also dictates impractically long
data records $N$ for reliable estimation. One approach to coping
with this dimensionality issue it to view polynomial modeling as a
kernel regression problem \cite{FrSc06}, \cite{ScSm02},
\cite{HaTiFr09}.

However, various applications admit sparse polynomial expansions,
where only a few, say $s$ out of $M$, expansion coefficients are
nonzero -- a fact that cannot be exploited via polynomial kernel
regression. The nonlinearity order, the memory size, and the nonzero
coefficients may all be unknown. Nonetheless, the polynomial
expansion in such applications is sparse -- an attribute that can be
due to either a parsimonious underlying physical system, or an
over-parameterized model assumed. Sparsity in polynomial expansions
constitutes the motivation behind this work. Volterra system
identification and polynomial regression are formulated in Section
\ref{sec:problem}. After explaining the link between the two
problems, several motivating applications with inherent sparse
polynomial structure are provided.

Section \ref{sec:estimation} deals with the estimation of sparse
polynomial expansions. Traditional polynomial filtering approaches
either drop the contribution of expansion terms \emph{a fortiori},
or adopt the sparsity-agnostic LS estimator \cite{MaSi00}.
Alternative estimators rely on: estimating a frequency-domain
equivalent model; modeling the nonlinear filter as the convolution
of two or more linear filters; transforming the polynomial
representation to a more parsimonious one (e.g., using the Laguerre
expansion); or by estimating fewer coefficients and then linearly
interpolating the full model; see \cite{MaSi00} and references
thereoff. However, the recent advances on compressive sampling
\cite{ChDoSa99}, \cite{CaTa05}, and the least-absolute shrinkage and
selection operator (Lasso) \cite{Tib96} offer a precious toolbox for
estimating sparse signals. Sparse Volterra channel estimators are
proposed in \cite{KeAnGia09} and \cite{MiBaKaTa10}. Building on
well-established (weighted) Lasso estimators \cite{Tib96},
\cite{Zou06}, and their efficient coordinate descent implementation
\cite{FrHaHoTi07}, the present paper develops an adaptive RLS-type
sparse polynomial estimation algorithm, which generalizes
\cite{AnBaGia10} to the nonlinear case, and constitutes the first
contribution.

Performance of the (weighted) Lasso estimators has been analyzed
asymptotically in the number of measurements $N$ \cite{FaLi01},
\cite{Zou06}. With finite samples, identifiability of Lasso-based
estimators and other compressive sampling reconstruction methods can
be assessed via the so-called restricted isometry properties (RIP)
of the involved regression matrix \cite{CaTa05}, \cite{BiRiTs09}. It
has been shown that certain random matrix ensembles satisfy
desirable properties with high probability when $N$ scales at least
as $s\log (M/s)$ \cite{CaTa05}. For Gaussian, Bernoulli, and uniform
Toeplitz matrices appearing in sparse linear filtering, the lower
bound on $N$ has been shown to scale as $s^2\log M$ \cite{HaNo10},
\cite{Ra10}. Section \ref{subsec:rip_volterra} deals with RIP
analysis for Volterra filters, which is the second contribution of
this work. It is shown that for a uniformly distributed input, the
second-order Volterra filtering matrix satisfies the RIP with high
probability when $N$ scales as $s^2\log M$, which extends the bound
from the linear to the Volterra filtering case.

The third contribution is the RIP analysis for the sparse polynomial
regression setup (Section \ref{subsec:rip_poly}). Because there are
no dependencies across rows of the involved regression matrix,
different tools are utilized and the resultant RIP bounds are
stronger than their Volterra filter counterparts. It is proved that
for a uniform input, $s$-sparse linear-quadratic regression requires
a number of measurements that scales as $s\log^4 L$. The same result
holds also for a model oftentimes employed for GWA analysis.

Applicability of the existing batch sparse estimators and their
developed adaptive counterparts is demonstrated through numerical
tests in Section \ref{sec:simulations}. Simulations on synthetic and
real GWA data show that sparsity-aware polynomial estimators can
cope with the curse of dimensionality and yield parsimonious yet
accurate models with relatively short data records. The work is
concluded in Section \ref{sec:conclusions}.

\emph{Notation:} Lower-(upper-)case boldface letters are reserved
for column vectors (matrices), and calligraphic letters for sets;
$\mathbf{1}_N$ denotes the all-ones vector of length $N$;
$(\cdot)^T$ denotes transposition;
$\mathcal{N}(\mathbf{m},\mathbf{\Sigma})$ stands for the
multivariate Gaussian probability density with mean $\mathbf{m}$ and
covariance matrix $\mathbf{\Sigma}$; $\mathbb{E}[\cdot]$ denotes the
expectation operator; $\|\mathbf{x}\|_p:=\left(\sum_{i=1}^n
|x_i|^p\right)^{1/p}$ for $p\geq 1$ stands for the $\ell_p$-norm in
$\mathbb{R}^n$, and $\|\mathbf{x}\|_0$ the $\ell_0$-(pseudo)norm,
which equals the number of nonzero entries of $\mathbf{x}$.

\section{Problem Formulation: Context and Motivation}\label{sec:problem}
Nonlinear system modeling using the Volterra expansion as well as
the more general notion of (multivariate) polynomial regression are
reviewed in this section. For both problems, the nonlinear I/O
dependency is expressed in the standard (linear with respect to the
unknown coefficients) matrix-vector form. After recognizing the
``curse of dimensionality'' inherent to the involved estimation
problems, motivating applications admitting (approximately)
\emph{sparse} polynomial representations are highlighted.

\subsection{Volterra Filter Model}\label{subsec:volterra}
Consider a nonlinear, discrete-time, and time-invariant I/O
relationship $y(n)=f\left( x(n),\ldots,x(1)\right)$, where $x(n)$
and $y(n)$ denote the input and output samples at time $n$. While
such nonlinear mappings can have infinite memory, finite-memory
truncation is adopted in practice to yield
$y(n)=f\left(\mathbf{x}_1(n)\right)$, where
$\mathbf{x}_1(n):=\left[x(n)~\ldots~x(n-L+1)\right]^T$ with $L$
finite. Under smoothness conditions, this I/O relationship can be
approximated by a Volterra expansion oftentimes truncated to a
finite order $P$ as
\begin{equation}\label{eq:expansion}
y(n)=\sum_{p=0}^{P}H_p\left[ \mathbf{x}_1(n)\right] +v(n)
\end{equation}
where $v(n)$ captures unmodeled dynamics and observation noise,
assumed to be zero-mean and independent of $\mathbf{x}_1(n)$ as well
as across time; and $H_p\left[ \mathbf{x}_1(n)\right]$ denotes the
output of the so-termed $p$-th order Volterra module
$h_p(k_1,\ldots,k_p)$ given by
\begin{equation}\label{eq:kernel}
H_p\left[ \mathbf{x}_1(n)\right]:=\sum_{k_1=0}^{L-1}
\ldots\sum_{k_p=0}^{L-1} h_p(k_1,\ldots,k_p) \prod_{i=1}^p x(n-k_i)
\end{equation}
where memory $L$ has been considered identical for all modules
without loss of generality. The Volterra expansion in
\eqref{eq:expansion}-\eqref{eq:kernel} has been thoroughly studied
in its representation power and convergence properties; see e.g.,
\cite{PaPo77}, \cite{MaSi00}, and references therein.

The goal here is to estimate $h_p(k_1,\ldots,k_p)$ for
$p=0,1,\ldots,P$, and $k_i=0,1,\ldots,L-1$, given the I/O samples
$\left\{\mathbf{x}_1(n),y(n)\right\}_{n=1}^N$, and upper bounds on
the expansion order $P$ and the memory size $L$. Although this
problem has been extensively investigated \cite{MaSi00}, the
sparsity present in the Volterra representation of many nonlinear
systems will be exploited here to develop efficient estimators.

To this end, \eqref{eq:expansion} will be expressed first in a
standard matrix-vector form \cite{MaSi00}. Define the vectors
$\mathbf{x}_p(n):=\mathbf{x}_{p-1}(n)\otimes \mathbf{x}_1(n)$ for
$p\geq 2$, where $\otimes$ denotes the Kronecker product; and write
the $p$-th order Volterra output as
$H_p\left[\mathbf{x}_1(n)\right]=\mathbf{x}_p^T(n)\mathbf{h}_p$,
where $\mathbf{h}_p$ contains the coefficients of
$h_p(k_1,\ldots,k_p)$ arranged accordingly. Using the latter,
\eqref{eq:expansion} can be rewritten as
\begin{equation}\label{eq:output_n}
y(n)=\mathbf{x}^T(n)\mathbf{h} + v(n),~n=1,\ldots,N
\end{equation}
where $\mathbf{x}(n):=\left[1 ~ \mathbf{x}_1^T(n) ~ \ldots ~ \mathbf{x}^T_P(n) \right]^T$, and
$\mathbf{h}:=\left[h_0 ~ \mathbf{h}_1^T ~ \ldots ~ \mathbf{h}_P^T \right]^T$. Concatenating \eqref{eq:expansion} for all $n$, one arrives at the linear model
\begin{equation}\label{eq:model}
\mathbf{y}=\mathbf{X}\mathbf{h} + \mathbf{v}
\end{equation}
where $\mathbf{y}:=\left[y(1)~\cdots~y(N)\right]^T$,
$\mathbf{X}:=\left[\mathbf{x}(1) ~ \ldots ~\mathbf{x}(N)\right]^T$,
and $\mathbf{v}:=\left[v(1)~\ldots~v(N)\right]^T$.

\subsection{Polynomial Regression Model}\label{subsec:polyreg}
Generalizing the Volterra filter expansion, polynomial regression
aims at approximating a nonlinear function $y(n)=f\left(
\{x_l(n)\}_{l=0}^{L-1}{}\right)$ of $L$ variables through an
expansion similar to \eqref{eq:expansion}-\eqref{eq:kernel}, where
the input vector $\mathbf{x}_1(n)$ is now defined as
$\mathbf{x}_1(n):=\left[x_0(n)~\ldots~x_{L-1}(n)\right]^T$, and $n$
is not necessarily a time index. Again the goal is to estimate
$h_p(k_1,\ldots,k_p)$ given $\{\mathbf{x}_1(n),y(n)\}_{n=1}^N$.
Polynomial regression can be interpreted as the $P$-th order Taylor
series expansion of $f\left(\mathbf{x}_1(n)\right)$, and appears in
several multilinear estimation and prediction problems in
engineering, natural sciences, and economics \cite{HaTiFr09}.

By simply choosing $x_l(n)=x(n-l)$ for $l=0,\ldots,L-1$, the
Volterra filter is a special case of polynomial regression. Since
this extra property has not been exploited in deriving
\eqref{eq:expansion}-\eqref{eq:model}, these equations carry over to
the polynomial regression setup. For this reason, the same notation
will be used henceforth for the two setups; the ambiguity will be
easily resolved by the context.

\subsection{The Curse of Dimensionality}\label{susec:curse}
Estimating the unknown coefficients in both the Volterra system
identification and in polynomial regression is critically challenged
by the curse of dimensionality. The Kronecker product defining
$\mathbf{x}_p(n)$ imply that the dimension of $\mathbf{h}_p$ is
$L^p$, and consequently $\mathbf{h}$ and $\mathbf{x}(n)$ have
dimension $\sum_{p=0}^P
L^p=\left(L^{P+1}-1\right)/\left(L-1\right)$. Note that all possible
permutations of the indices $\{k_1,\ldots,k_p\}$ multiply the same
input term $x_{k_1}(n)\cdots x_{k_p}(n)$; e.g., $h_2(0,1)$ and
$h_2(1,0)$ both multiply the monomial $x_0(n)x_1(n)$. To obtain a
unique representation of \eqref{eq:kernel}, only one of these
permutations is retained. After discarding the redundant
coefficients, the dimension of $\mathbf{h}_p$ and
$\mathbf{x}_p(n)$'s is reduced to $\binom{L+p-1}{p}$ \cite{MaSi00}.
Exploiting such redundancies in modules of all orders eventually
shortens $\mathbf{h}$ and $\mathbf{x}(n)$'s to dimension
\begin{equation}\label{eq:dimension}
M:=\sum_{p=0}^{P}\binom{L+p-1}{p}=\binom{L+P}{P}=\binom{L+P}{L}
\end{equation}
which still grows fast with increasing $L$ and $P$. For notational
brevity, $\mathbf{h}$ and $\mathbf{X}$ will denote the shortened
versions of the variables in \eqref{eq:model}; that is matrix
$\mathbf{X}$ will be $N\times M$.

\subsection{Motivating Applications}\label{subsec:sparse_poly_exp}
Applications are outlined here involving models that admit
(approximately) sparse polynomial representations. When $P$ and $L$
are unknown, model order selection can be accomplished via
sparsity-cognizant estimators. Beyond this rather mundane task,
sparsity can arise due to problem specifications, or be imposed for
interpretability purposes.

A special yet widely employed Volterra model is the so-called
linear-nonlinear-linear (LNL) one \cite{MaSi00}. It consists of a
linear filter with impulse response $\{h_a(k)\}_{k=0}^{L_a-1}$, in
cascade with a memoryless nonlinearity $f(x)$, and a second linear
filter $\{h_b(k)\}_{k=0}^{L_b-1}$. The overall memory is thus
$L=L_a+L_b-1$. If $f(x)$ is analytic on an open set $(a,b)$, it
accepts a Taylor series expansion $f(x)=\sum_{p=0}^{\infty}c_p x^p$
in $x\in (a,b)$. It can be shown that the $p$-th order redundant
Volterra module is given by \cite[Ch.~2]{MaSi00}
\begin{equation}\label{eq:LNL}
\hspace*{-0.4em}h_p(k_1,\ldots,k_p)=c_p\sum_{k=0}^{L_b-1}h_b(k)h_a(k_1-k)\ldots h_a(k_p-k)
\end{equation}
for $k_i\in \{0,\ldots,L-1\}$. In \eqref{eq:LNL}, there are
$p$-tuples $(k_1,\ldots,k_p)$ for which there is no $k\in
\{0,\ldots,L_b-1\}$ such that $(k_i-k)\in \{0,\ldots,L_a-1\}$ for
all $i=1,\ldots,p$. For these $p$-tuples, the corresponding Volterra
coefficient is zero. As an example, for filters of length
$L_a=L_b=6$ and for $P=3$, among the $364$ non-redundant Volterra
coefficients, the nonzero ones are no more than $224$. When $L_a$
and $L_b$ are not known, the locations of the zero coefficients
cannot be determined a priori. By dropping the second linear filter
in the LNL model, the Wiener model is obtained. Its Volterra modules
follow immediately from \eqref{eq:LNL} and have the separable form
$h_p(k_1,\ldots,k_p)=c_ph_a(k_1)\ldots h_a(k_p)$ for every $p$
\cite{MaSi00}. Likewise, by ignoring the first filter, the LNL model
is transformed to the so-called Hammerstein model in which
$h_p(k_1,\ldots,k_p)=c_ph_b(k)$ for $k=k_1=\ldots=k_p$; and 0
otherwise. The key observation in all three models is that if at
least one of the linear filters is sparse, the resulting Volterra
filter is even sparser.

That is usually the case when modeling the nonlinear behavior of
loudspeakers and high-power amplifiers (HPA) \cite{MaSi00},
\cite{BeBi83}. When a small-size (low-cost) loudspeaker is located
close to a microphone (as is the case in cellular phones,
teleconferencing, hands-free, or hearing aid systems), the
loudspeaker sound is echoed by the environment before arriving at
the microphone. A nonlinear acoustic echo canceller should
adaptively identify the impulse response comprising the loudspeaker
and the room, and thereby subtract undesirable echoes from the
microphone signal. The cascade of the loudspeaker, typically
characterized by a short memory LNL or a Wiener model, and the
typically long but (approximately) sparse room impulse response
gives rise to a sparse Volterra filter \cite{ZeKe10}. Similarly,
HPAs residing at the transmitters of wireless communication links
are usually modeled as LNL structures having only a few coefficients
contributing substantially to the output \cite[p.60]{BeBi83}. When
the HPA is followed by a multipath wireless channel represented by a
sparse impulse response, the overall system becomes sparse too
\cite{MiBaKaTa10}.

Sparse polynomial expansions are also encountered in neuroscience
and bioinformatics. Volterra filters have been adopted to model
causal relationships in neuronal ensembles using spike-train data
recorded from individual neurons \cite{Mar10}, \cite{SoWaBe10}.
Casting the problem as a probit Volterra regression, conventional
model selection techniques have been pursued to zero blocks of
Volterra expansion coefficients, and thus reveal neuron connections.
Furthermore, genome-wide association (GWA) analysis depends
critically on sparse polynomial regression models \cite{Co09},
\cite{WuCh09}, \cite{Xu07}. Through GWA studies, geneticists
identify which genes determine certain phenotypes, e.g., human
genetic diseases or traits in other species. Analysis has revealed
that genetic factors involve multiplicative interactions among genes
-- a fact known as epistasis; hence, linear gene-phenotype models
are inadequate. The occurrence of a disease can be posed as a
(logistic) multilinear regression, where apart from single-gene
terms, the output depends on products of two or more genes as well
\cite{Co09}. To cope with the under-determinacy of the problem and
detect gene-gene interactions, sparsity-promoting logistic
regression methods have been developed; see e.g., \cite{WuCh09}.

Based on these considerations, exploiting sparsity in polynomial
representations is well motivated and prompted us to develop the
sparsity-aware estimators described in the following section.

\section{Estimation of Sparse Polynomial Expansions}\label{sec:estimation}
One of the attractive properties of Volterra and polynomial
regression models is that the output is a linear function of the
wanted coefficients. This allows one to develop standard estimators
for $\mathbf{h}$ in \eqref{eq:model}. However, the number of
coefficients $M$ can be prohibitively large for reasonable values of
$P$ and $L$, even after removing redundancies. Hence, accurately
estimating $\mathbf{h}$ requires a large number of measurements $N$
which: i) may be impractical and/or violate the stationarity
assumption in an adaptive system identification setup; ii) entails
considerable computational burden; and iii) raises numerical
instability issues. To combat this curse of dimensionality, batch
\emph{sparsity-aware} methods will be proposed first for polynomial
modeling, and based on them, adaptive algorithms will be developed
afterwards.

\subsection{Batch Estimators}\label{subsec:batch}
Ignoring $\mathbf{v}$ in \eqref{eq:model}, the vector $\mathbf{h}$
can be recovered by solving the linear system of equations
$\mathbf{y}=\mathbf{X}\mathbf{h}$. Generally, a unique solution is
readily found if $N\geq M$; but when $N<M$, there are infinitely
many solutions. Capitalizing on the sparsity of $\mathbf{h}$, one
should ideally solve
\begin{equation}\label{eq:ell0}
\min_{\mathbf{h}} ~ \left\{\|\mathbf{h}\|_0:~ \mathbf{y}=\mathbf{X}\mathbf{h}\right\}.
\end{equation}
Recognizing the NP-hardness of solving \eqref{eq:ell0}, compressive
sampling suggests solving instead the linear program
\cite{ChDoSa99}, \cite{CaTa05}
\begin{equation}\label{eq:bp}
\min_{\mathbf{h}} ~ \left\{\|\mathbf{h}\|_1:~ \mathbf{y}=\mathbf{X}\mathbf{h}\right\}
\end{equation}
which is also known as basis pursuit and can quantifiably
approximate the solution of \eqref{eq:ell0}; see Section
\ref{sec:rip} for more on the relation between \eqref{eq:ell0} and
\eqref{eq:bp}. However, modeling errors and measurement noise,
motivate a LS estimator
$\hat{\mathbf{h}}^{LS}:=\arg\min_{\mathbf{h}}
\|\mathbf{y}-\mathbf{X}\mathbf{h}\|_2^2$. If $N\geq M$ and
$\mathbf{X}$ has full column rank, the LS solution is uniquely found
as $\hat{\mathbf{h}}^{LS} = \left( \mathbf{X}^T \mathbf{X}
\right)^{-1}\mathbf{X}^T\mathbf{y}$. If the input is drawn either
from a continuous distribution or from a finite alphabet of at least
$P{+}1$ values, $\mathbf{X}^T\mathbf{X}$ is invertible almost
surely; but its condition number grows with $L$ and $P$
\cite{NoVeen95}. A large condition number translates to numerically
ill-posed inversion of $\mathbf{X}^T\mathbf{X}$ and amplifies noise
too. If $N<M$, the LS solution is not unique; but one can choose the
minimum $\ell_2$-norm solution $\hat{\mathbf{h}}^{LS} =\mathbf{X}^T
\left( \mathbf{X} \mathbf{X}^T \right)^{-1}\mathbf{y}$.

For both over/under-determined cases, one may resort to the ridge
($\ell_2$-norm regularized) solution
\begin{subequations}\label{eq:ridge}
\begin{align}
\hat{\mathbf{h}}^{Ridge}&:= \left(\mathbf{X}^T\mathbf{X} +
\delta\mathbf{I}_{M}\right)^{-1}\mathbf{X}^T\mathbf{y}\label{eq:ridge1}\\
& = \mathbf{X}^T\left(\mathbf{X}\mathbf{X}^T +
\delta\mathbf{I}_{N}\right)^{-1}\mathbf{y}\label{eq:ridge2}
\end{align}
\end{subequations}
for some $\delta>0$, where the equality can be readily proved by
algebraic manipulations. Calculating, storing in the main memory,
and inverting the matrices in parentheses are the main bottlenecks
in computing $\hat{\mathbf{h}}^{Ridge}$ via \eqref{eq:ridge}.
Choosing \eqref{eq:ridge1} versus \eqref{eq:ridge2} depends on how
$N$ and $M$ compare. Especially for polynomial (or Volterra)
regression, the $(n_1,n_2)$-th entry of $\mathbf{X}\mathbf{X}^T$,
which is the inner product $\mathbf{x}^T(n_1)\mathbf{x}(n_2)$, can
be also expressed as $\sum_{p=0}^P
\left(\mathbf{x}^T_{1}(n_1)\mathbf{x}_{1}(n_2)\right)^p$. This
computational alternative is an instantiation of the so-called
kernel trick, and reduces the cost of computing
$\mathbf{X}\mathbf{X}^T$ in \eqref{eq:ridge2} from
$\mathcal{O}(N^2M)$ to $\mathcal{O}(N^2(L+P))$ \cite{ScSm02},
\cite{FrSc06}; see also Subsection \ref{subsec:poly_kernel}.

In any case, neither $\hat{\mathbf{h}}^{LS}$ nor
$\hat{\mathbf{h}}^{Ridge}$ are sparse. To effect sparsity, the idea
is to adopt as regularization penalty the $\ell_1$-norm of the
wanted vector \cite{Tib96}
\begin{equation}\label{eq:Lasso}
\hat{\mathbf{h}}=\arg\min_{\mathbf{h}} \frac{1}{2}\|\mathbf{y}-
\mathbf{X}\mathbf{h}\|_2^2+\lambda_N\sum_{i=1}^M w_i |h_i|
\end{equation}
where $h_i$ is the $i$-th entry of $\mathbf{h}$, and $w_i>0$ for $i=1,\ldots,M$.
Two choices of $w_i$ are commonly adopted:
\begin{description}
\item[(w1)] $w_i=1$ for $i=1,\ldots,M$, which corresponds to the
conventional Lasso estimator \cite{Tib96}; or,
\item[(w2)] $w_i=|\hat{h}^{Ridge}_i|^{-1}$ for $i=1,\ldots,M$, which
leads to the weighted Lasso estimator \cite{Zou06}.
\end{description}
Asymptotic performance of the Lasso estimator has been analyzed in
\cite{FaLi01}, where it is shown that the weighted Lasso estimator
exhibits improved asymptotic properties over Lasso at the price of
requiring the ridge regression estimates to evaluate the $w_i$'s
\cite{Zou06}. For the practical finite-sample regime, performance of
the Lasso estimator is analyzed through the restricted isometry
properties of $\mathbf{X}$ in Section \ref{sec:rip}, where rules of
thumb are also provided for the selection of $\lambda_N$ as well
(cf. Lemma \ref{le:Lasso_RIP}).

Albeit known for \emph{linear} regression models, the novelty here
is the adoption of (weighted) Lasso for \emph{sparse polynomial}
regressions. Sparse generalized linear regression models, such as
$\ell_1$-regularized logistic and probit regressions can be fit as a
series of successive Lasso problems after appropriately redefining
the response $\mathbf{y}$ and weighting the input $\mathbf{X}$
\cite[Sec.~4.4.1]{HaTiFr09}, \cite{WuCh09}. Hence, solving and
analyzing Lasso for sparse polynomial expansions is important for
generalized polynomial regression as well. Moreover, in certain
applications, Volterra coefficients are collected in subsets
(according to their order or other criteria) that are effected to be
(non)zero as a group \cite{SoWaBe10}. In such applications, using
methods promoting group-sparsity is expected to improve
recoverability \cite{YuLi06}. Even though sparsity is manifested
here at the single-coefficient level, extensions toward the
aforementioned direction constitutes an interesting future research
topic.

Algorithmically, the convex optimization problem in \eqref{eq:Lasso}
can be tackled by any generic second-order cone program (SOCP)
solver, or any other method tailored for the Lasso estimator. The
method of choice here is the coordinate descent scheme of
\cite{FrHaHoTi07}, which is outlined next for completeness. The core
idea is to iteratively minimize \eqref{eq:Lasso} w.r.t. one entry of
$\mathbf{h}$ at a time, while keeping the remaining ones fixed, by
solving the scalar minimization problem
\begin{equation}\label{eq:Lassoi}
\min_{h_i} \frac{1}{2}\|\mathbf{y}-
\mathbf{X}^{(-i)}\hat{\mathbf{h}}^{(-i)}-\mathbf{x}_i h_i\|_2^2+\lambda_N w_i |h_i|
\end{equation}
where $\mathbf{x}_i$ is the $i$-th column\footnote{Recall that
$\mathbf{x}(n)$ stands for the $n$-th row of $\mathbf{X}$.} of
$\mathbf{X}$, variables $\mathbf{X}^{(-i)}$  and
$\hat{\mathbf{h}}^{(-i)}$ denote $\mathbf{X}$ and
$\hat{\mathbf{h}}$, respectively, having the $i$-th column (entry)
removed, and $\hat{\mathbf{h}}$ is the latest value for the optimum
$\mathbf{h}$. It turns out that the component-wise minimization of
\eqref{eq:Lassoi} admits the closed-form solution \cite{FrHaHoTi07}
\begin{equation}\label{eq:hi_formula}
\hat{h}_i\leftarrow \frac{\sign(z_i)}{R_{ii}}\cdot \left[|z_i|-\lambda_N w_i\right]_+
\end{equation}
where $\left[x\right]_+:=\max(x,0)$, $R_{ii}$ is the $i$-th entry of
the sample correlation or Grammian matrix
$\mathbf{R}:=\mathbf{X}^T\mathbf{X}$ and $z_i$ is the $i$-th entry
of $\mathbf{z}_i:=\mathbf{X}^T\left( \mathbf{y} -
\mathbf{X}^{(-i)}\hat{\mathbf{h}}^{(-i)} \right)$. After
initializing $\hat{\mathbf{h}}$ to any value (usually zero), the
algorithm iterates by simply updating the entries of
$\hat{\mathbf{h}}$ via \eqref{eq:hi_formula}. By defining
$\mathbf{z}:=\mathbf{X}^T\left( \mathbf{y} -
\mathbf{X}\hat{\mathbf{h}}\right)$, vector $\mathbf{z}_i$ can be
updated as
\begin{equation}\label{eq:update1}
\mathbf{z}_i\leftarrow\mathbf{z} + \mathbf{r}_i \hat{h}_i
\end{equation}
with $\mathbf{r}_i$ being the $i$-th column of $\mathbf{R}$. After
updating $\hat{h}_i$ to its new value \eqref{eq:hi_formula},
$\mathbf{z}$ has to be updated too as
\begin{equation}\label{eq:update2}
\mathbf{z}\leftarrow\mathbf{z}_i - \mathbf{r}_i \hat{h}_i.
\end{equation}
It is easy to see that $\{\mathbf{z}_i\}_{i=1}^M$ in
\eqref{eq:update1}-\eqref{eq:update2} are not essentially needed,
and one can update only $\mathbf{z}$. These iterates constitute the
cyclic coordinate descent (CCD) algorithm for the (weighted) Lasso
problem, and are tabulated as Alg.~\ref{alg:batch}. CCD-(W)L is
guaranteed to converge to a minimizer of \eqref{eq:Lasso}
\cite{FrHaHoTi07}. Apart from the initial computation of
$\mathbf{z}$ and $\mathbf{R}$ which incurs complexity
$\mathcal{O}(M^2N)$, the complexity of Alg.~\ref{alg:batch} as
presented here is $\mathcal{O}(M)$ per coordinate iteration; see
also \cite{FrHaHoTi07}.

\subsection{Recursive Estimators}\label{subsec:sequential}
Unlike batch estimators, their recursive counterparts offer
computational and memory savings, and enable tracking of slowly
time-varying systems. The recursive LS (RLS) algorithm is an
efficient implementation of the LS, and the ridge estimators. It
solves sequentially the following problem:
\begin{equation}\label{eq:RLS-opt}
\hat{\mathbf{h}}^{RLS}_N:=\arg\min_{\mathbf{h}}\sum_{n=1}^N\beta^{N-n}\left(y(n)-\mathbf{x}^T(n)
\mathbf{h}\right)^2 +\beta^{N}\delta\|\mathbf{h}\|_2^2
\end{equation}
where $\beta$ denotes the forgetting factor and $\delta$ a small
positive constant. For time-invariant systems, $\beta$ is set to
$1$, while $0\ll \beta<1$ enables tracking of slow variations.
Similar to the batch LS, the RLS does not exploit the a priori
knowledge on the sparsity of $\mathbf{h}$, and suffers from
numerical instability especially when the effective memory of the
algorithm, $1/(1-\beta)$, is comparable to the dimension $M$ of
$\mathbf{h}$.

To overcome these limitations, the following approach is advocated
for polynomial regression:
\begin{equation}\label{eq:RLasso}
\hat{\mathbf{h}}_N=\arg\min_{\mathbf{h}} J^{L}_N(\mathbf{h})
\end{equation}
\begin{equation*}
J^{L}_N(\mathbf{h}):=\sum_{n=1}^N\beta^{N-n}\left(y(n)-\mathbf{x}^T(n)
\mathbf{h}\right)^2 +\lambda_{N}\sum_{i=1}^M w_{N,i}|h_i|
\end{equation*}
where $w_{N,i}$ can be chosen as
\begin{description}
\item[(a1)] $w_{N,i}=1$ $\forall N$, $i=1,\ldots,M$, which corresponds to the
recursive Lasso (RL) problem; or,
\item[(a2)] $w_{N,i}=|\hat{h}^{RLS}_{N,i}|^{-1}$ $\forall N$, $i=1,\ldots,M$,
leading to the recursive weighted Lasso (RWL) one.
\end{description}
The sequence $\{\hat{\mathbf{h}}_N\}$ cannot be updated recursively,
and \eqref{eq:RLasso} calls for a convex optimization solver for
each time instant or measurement $N$. To avoid the computational
burden involved, several methods have been developed for sparse
linear models; see \cite{AnBaGia10} and the references therein. The
coordinate descent algorithm of Subsection \ref{subsec:batch} can be
extended to \eqref{eq:RLasso} by first updating $\mathbf{R}$ and
$\mathbf{z}$ as
\begin{subequations}\label{eq:Rz_updates}
\begin{align}
\mathbf{R}_N&=\beta\mathbf{R}_{N-1}+\mathbf{x}(N)\mathbf{x}^T(N)\label{eq:R_update}\\
\mathbf{z}_N&=\beta\mathbf{z}_{N-1}+\mathbf{x}(N)(y(N)-\mathbf{x}^T(N)\hat{\mathbf{h}}_{N-1})\label{eq:z_update}
\end{align}
\end{subequations}
where $\hat{\mathbf{h}}_{N-1}$ is a solution at time $N-1$. The
minimizer $\hat{\mathbf{h}}_N$ can then be found by performing
component-wise minimizations until convergence in the spirit of the
corresponding batch estimator. However, to speed up computations and
leverage the adaptivity of the solution, we choose to perform a
single cycle of component-wise updates. Thus, $\hat{\mathbf{h}}_N$
is formed by the iterates of the inner loop in
Alg.~\ref{alg:recursive}, where $\mathbf{r}_{N,i}$, $z_{N,i}$,
$\mathbf{R}_{N,ii}$, and $\hat{h}_{N,i}$ are defined as before.


The presented algorithm called hereafter cyclic coordinate descent
for recursive (weighted) Lasso (CCD-R(W)L) is summarized as
Alg.~\ref{alg:recursive}; the convergence properties of CCD-RL have
been established in \cite{AnBaGia10} for linear regression, but
carry over directly to the polynomial regression considered here.
Its complexity is $\mathcal{O}(M^2)$ per measurement which is of the
same order as the RLS. By setting $w_{N,i}=0$ or
$w_{N,i}=|\hat{h}^{RLS}_{N,i}|^{-1}$, the CCD-R(W)L algorithms
approximate the minimizers of the R(W)L problems.

\subsection{Polynomial Reproducing Kernels}\label{subsec:poly_kernel}
An alternative approach to polynomial modeling is via kernel
regression \cite{ScSm02}. In the general setup, kernel regression
approximates a nonlinear function $f(\mathbf{x}_1)$ assuming it can
be linearly expanded over a possibly infinite number of basis
functions $\phi_k(\mathbf{x}_1)$ as
$f(\mathbf{x}_1)=\sum_{k=1}^{K}\alpha_k\phi_k(\mathbf{x}_1)$. When
$\phi_k(\mathbf{x}_1)=\kappa\left(\mathbf{x}_1,\mathbf{x}_1(k)\right)$
with $\kappa(\cdot,\cdot)$ denoting a judiciously selected positive
definite kernel, $f(\mathbf{x}_1)$ lies in a reproducing kernel
Hilbert space $\mathcal{H}$, and kernel regression is formulated as
the variational problem
\begin{equation}\label{eq:kernel_regression}
\min_{f\in\mathcal{H}} C\left(\{f\left(\mathbf{x}_1(n)\right),y(n)\}_{n=1}^N\right) + \|f\|_{\mathcal{H}}
\end{equation}
where $C(\cdot)$ is an arbitrary cost function, and
$\|f\|_{\mathcal{H}}$ is the norm in $\mathcal{H}$ that penalizes
complexity of $f$. It turns out that there exists a minimizer of
\eqref{eq:kernel_regression} expressed as
$f(\mathbf{x}_1)=\sum_{n=1}^N\alpha_n
\kappa\left(\mathbf{x}_1,\mathbf{x}_1(n)\right)$, while for many
meaningful costs the $\alpha_n$'s can be computed in
$\mathcal{O}(N^3)$ using convex optimization solvers \cite{ScSm02}.

Polynomial regression can be cast as kernel regression after setting
$\kappa(\mathbf{x}_1(n_1),\mathbf{x}_1(n_2))$ to be either the
homogeneous polynomial kernel
$\left(\mathbf{x}^T_1(n_1)\mathbf{x}_1(n_2)\right)^P$, or, one of
the inhomogeneous ones
$\left(1{+}\mathbf{x}^T_1(n_1)\mathbf{x}_1(n_2)\right)^P$ or
$\sum_{p=0}^P\left(\mathbf{x}^T_1(n_1)\mathbf{x}_1(n_2)\right)^p$
\cite{ScSm02}, \cite{FrSc06}. Once the $\alpha_n$'s have been
estimated, the polynomial coefficients $\mathbf{h}$ (cf.
\eqref{eq:model}) can be found in closed form \cite{FrSc06}.
Furthermore, objectives $C(\cdot)$ such as the
$\epsilon$-insensitive cost, yield sparsity in the
$\alpha_n$--domain, and thus designate the so-called support vectors
among the $\mathbf{x}_1(n)$'s \cite{ScSm02}. Even though kernel
regression alleviates complexity concerns, the $\mathbf{h}$ which
can indirectly obtained cannot be sparse. Thus, sparsity-aware
estimation in the primal $\mathbf{h}$--domain (as opposed to the
dual $\alpha_n$--domain) comes with interpretational and modeling
advantages.

\section{Identifiability of Sparse Polynomial Models}\label{sec:rip}
This section focuses on specifying whether the optimization problems
in \eqref{eq:bp} and \eqref{eq:Lasso} are capable of identifying a
sparse polynomial expansion. The asymptotic in $N$ behavior of the
(weighted) Lasso estimator has been studied in \cite{FaLi01},
\cite{Zou06}; practically though one is more interested in
finite-sample recoverability guarantees. One of the tools utilized
to this end is the so-called \textit{restricted isometry properties}
(RIP) of the involved regression matrix $\mathbf{X}$. These are
defined as \cite{CaTa05}:
\begin{definition}[Restricted Isometry Properties (RIP)]\label{def:RIP}
Matrix $\mathbf{X}\in\mathbb{R}^{N\times M}$ possesses the
restricted isometry of order $s$, denoted as $\delta_s\in (0,1)$, if
for all $\mathbf{h}\in\mathbb{R}^{M}$ with $\|\mathbf{h}\|_0\leq s$
\begin{equation}\label{eq:RIP}
\left(1-\delta_s\right)\|\mathbf{h}\|_2^2\leq \|\mathbf{Xh}\|_2^2\leq  \left(1+\delta_s\right)\|\mathbf{h}\|_2^2.
\end{equation}
\end{definition}

RIP were initially derived to provide identifiability conditions of
an $s$-sparse vector $\mathbf{h}_o$ given noiseless linear
measurements $\mathbf{y}=\mathbf{X}\mathbf{h}_o$. It has been shown
that the $\ell_0$-pseudonorm minimization in \eqref{eq:ell0} can
uniquely recover $\mathbf{h}_o$ if and only if $\delta_{2s}<1$. If
additionally $\delta_{2s}<\sqrt{2}-1$, then $\mathbf{h}_o$ is the
unique minimizer of the basis pursuit cost in \eqref{eq:bp}
\cite{Ca08}.

RIP-based analysis extends to noisy linear observations of an
$s$-sparse vector; that is, for  $\mathbf{y}=\mathbf{X} \mathbf{h}_o
+\mathbf{v}$. If $\|\mathbf{v}\|_2\leq \epsilon$, the constrained
version of the Lasso optimization problem
\begin{align}\label{eq:BN}
\min_{\mathbf{h}} ~ \left\{\|\mathbf{h}\|_1: ~ \|\mathbf{y}-\mathbf{X}\mathbf{h}\|_2\leq \epsilon \right\}
\end{align}
yields $\|\hat{\mathbf{h}}_{BN}-\mathbf{h}_o\|_2^2 \leq
c_{BN}^2\cdot\epsilon^2$, where
$c_{BN}:=\frac{4\left(1+\delta_{2s}\right)}{1-\delta_{2s}\left(\sqrt{2}+1\right)}$
whenever $\delta_{2s}<\sqrt{2}-1$ \cite{Ca08}. Furthermore, if
$\mathbf{v}\sim \mathcal{N}(\mathbf{0},\sigma^2\mathbf{I}_N)$, the
Dantzig selector defined as
\begin{align}\label{eq:Dantzig}
\min_{\mathbf{h}}~\left\{ \|\mathbf{h}\|_1: ~\|\mathbf{X}^T\left(\mathbf{y}- \mathbf{X}\mathbf{h}\right)\|_{\infty} \leq \epsilon_{DS} \right\}
\end{align}
satisfies $\|\hat{\mathbf{h}}_{DS}-\mathbf{h}_o\|_2^2 \leq
c_{DS}\cdot\sigma^2 s \log M$, where $c_{DS}:=\left(\frac{4\sqrt{2}}
{1-\delta_{2s}\left(\sqrt{2}+1\right)}\right)^2$ with probability at
least $1-\left( \pi \log M \right)^{-1/2}$ whenever
$\delta_{2s}<\sqrt{2}-1$, and $\epsilon_{DS}=\sqrt{2}\sigma
\sqrt{\log M}$ \cite{CaTa07}. Similarly, RIP-based recoverability
guarantees can be derived in the stochastic noise setting for the
Lasso estimator as described in the following lemma.

\begin{lemma}\label{le:Lasso_RIP}
Consider the linear model
$\mathbf{y}=\mathbf{X}\mathbf{h}_o+\mathbf{v}$, where the columns of
$\mathbf{X}\in \mathbb{R}^{N\times M}$ are of unit $\ell_2$-norm,
$\|\mathbf{h}_o\|_0=s$, and $\mathbf{v}\sim
\mathcal{N}(\mathbf{0},\sigma^2\mathbf{I}_N)$. Let
$\hat{\mathbf{h}}_L$ denote the minimizer of the Lasso estimator
\eqref{eq:Lasso} with $w_i=1$ for $i=1,\ldots,M$, and
$\lambda=A\sigma\sqrt{\log M}$ for $A>2\sqrt{2}$. If
$\delta_{2s}<\frac{1}{3\sqrt{2}+1}$, the bounds
\begin{align}
&\|\hat{\mathbf{h}}_L-\mathbf{h}_o\|_1\leq \frac{16A}{c_L}\cdot\sigma s\sqrt{\log M}\label{eq:bound_l1}\\
&\|\hat{\mathbf{h}}_L-\mathbf{h}_o\|_2^2\leq \left(\frac{16A}{c_L}\right)^2\cdot\sigma^2 s\log M\label{eq:bound_l2}\\
&\|\mathbf{X}(\hat{\mathbf{h}}_L-\mathbf{h}_o)\|_2^2\leq \frac{16A^2}{c_L}\cdot\sigma^2 s\log M\label{eq:bound_rec}
\end{align}
hold with probability at least $1-M^{1-A^2/8}$ for
$c_L=\left(1-\delta_{2s}\right)\left(1-
\frac{3\sqrt{2}\delta_{2s}}{1-\delta_{2s}}\right)^2$.
\end{lemma}

\begin{IEEEproof}
The lemma follows readily by properly adapting Lemma~4.1 and
Theorem~7.2 of \cite{BiRiTs09}.
\end{IEEEproof}

The earlier stated results document and quantify the role of
RIP-based analysis in establishing identifiability in a compressive
sampling setup. However, Definition \ref{def:RIP} suggests that
finding the RIP of a given matrix $\mathbf{X}$ is probably a hard
combinatorial problem. Thus, to derive sparse recoverability
guarantees one usually resorts to random matrix ensembles to provide
probabilistic bounds on their RIP \cite{CaTa05}, \cite{Ra10}. In the
generic sparse linear regression setup, it has been shown that when
the entries of $\mathbf{X}\in \mathbb{R}^{N\times M}$ are
independently Gaussian or Bernoulli, $\mathbf{X}$ possesses RIP
$\delta_s$ with probability at least
$1-\exp\left(-\delta_s^2/(2C)\right)$ when the number of
measurements is $N\geq 2C/\delta_s^2\cdot s\log(M/s)$, where $C$ is
a universal constant; this bound is known to be optimal
\cite{CaTa05}. In a sparse system identification setup where the
regression matrix has a Toeplitz structure, the condition on the
number of measurements $N$ obtained so far loosens to a scaling of
$s^2\log M$ for a Gaussian, Bernoulli, or uniform input
\cite{HaNo10}, \cite{Ra10}. The quadratic scaling of $N$ w.r.t. $s$
in the latter bound versus the linear scaling in the former can be
attributed to the statistical dependencies among the entries of
$\mathbf{X}$ \cite{Ra10}. Our contribution pertains to
characterizing the RIP of the involved regression matrix for both
the Volterra system identification and the multivariate polynomial
regression scenarios.

\subsection{RIP for Volterra System Identification}\label{subsec:rip_volterra}
For the Volterra filtering problem under study, the following
assumptions will be in force:
\begin{description}
\item[(as1)] input $\{x_n\}$ is independently drawn from the uniform distribution, i.e., $x_n \sim\mathcal{U}[-1,1]$; and
\item[(as2)] expansion is of order $P=2$ (linear-quadratic Volterra model).
\end{description}
Regarding (as1), recall that the Volterra expansion is a Taylor
series approximation of a nonlinear function; thus, it is reasonable
to focus on a bounded input region. Moreover, practically, one is
frequently interested in the behavior of a nonlinear system for a
limited input range. For (as2), the non-homo\-gene\-ous quadratic
Volterra model is a commonly adopted one. Generalization to models
with $P\geq 3$ is not straightforward and goes beyond the scope of
our RIP analysis. The considered Volterra filter length is
$M=\binom{L+2}{2}$; and, for future use, it is easy to check that
under (as1) it holds that $\mathbb{E}[x_n^2]=1/3$ and
$\mathbb{E}[x_n^4]=1/5$.

To start, recall the definition of the Grammian matrix
$\mathbf{R}:=\mathbf{X}^T\mathbf{X}$ and let $R_{ij}$ denote its
$(i,j)$-th entry. As shown in \cite[Sec.~III]{HaNo10}, the matrix
$\mathbf{X}$ possesses RIP $\delta_{s}$ if there exist positive
$\delta_d$ and $\delta_o$ with $\delta_d+\delta_o=\delta_s$ such
that  $|R_{ii}-1|<\delta_d$ and $|R_{ij}|<\delta_o/s$ for every
$i,j$ with $j\neq i$. When these conditions hold, Ger\v{s}gorin's
disc theorem guarantees that the eigenvalues of Grammian matrices
formed by any combination of $s$ columns of $\mathbf{X}$ lie in the
interval $[1-\delta_{s},1+\delta_{s}]$, and $\mathbf{X}$ possesses
RIP $\delta_{s}$ by definition. In a nutshell, for a regression
matrix $\mathbf{X}$ to have small $\delta_{s}$'s, and hence
favorable compressed sampling properties, it suffices that its
Grammian matrix has diagonal entries close to unity and off-diagonal
entries close to zero. If the involved regression matrix
$\mathbf{X}$ had unit $\ell_2$-norm columns, then the $\{R_{ii}\}$
would be unity by definition and one could merely study the quantity
$\max_{i,j,j\neq i} |R_{ij}|$, defined as the coherence of
$\mathbf{X}$; see also \cite[p.~13]{Ra10} for the relation between
coherence and the RIP.

In the Volterra filtering problem at hand, the diagonal entries
$\{R_{ii}\}$ are not equal to one; but an appropriate normalization
of the columns of $\mathbf{X}$ can provide at least
$\mathbb{E}[R_{ii}]=1$ for all $i$. The law of large numbers
dictates that given sufficiently enough measurements $N$, the
$R_{ii}$'s will approach their mean value. Likewise, it is desirable
for the off-diagonal entries of $\mathbf{R}$ to have zero mean, so
that they vanish for large $N$. Such a requirement is not inherently
satisfied by all $R_{ij}$'s with $j\neq i$; e.g., the inner product
between $\mathbf{X}$ columns of the form $\left[x_n^2 ~
x_{n+1}^2~\ldots~x_{n+N-1}^2 \right]^T$ and $\left[x_{n-k}^2 ~
x_{n-k+1}^2~\ldots~x_{n-k+N-1}^2 \right]^T$ for some $n$ and $k> 0$
has expected value $N\left(\mathbb{E}[x_n^2]\right)^2$ that is
strictly positive.

To achieve the desired properties, namely
\begin{description}
\item[(p1)] $\mathbb{E}[R_{ii}]=1$ for all $i=1,\ldots,M$, and
\item[(p2)] $\mathbb{E}[R_{ij}]=0$ for all $i,j=1,\ldots,M$ and $j\neq i$
\end{description}
it will be soon established that instead of studying the RIP of
$\mathbf{X}$, one can equivalently focus on its modified version
$\mathbft{X}\in \mathbb{R}^{N\times M}$ defined as
\begin{equation}\label{eq:Xt_def}
\mathbft{X}:=\left[\begin{array}{cccc}
\mathbft{x}^c & \mathbft{X}^{l} & \mathbft{X}^{q} & \mathbft{X}^{b}
\end{array}\right]
\end{equation}
where $\mathbft{x}^c:=\mathbf{1}_N/\sqrt{N}$ corresponds to the
constant (intercept or dc) component, $\mathbft{X}^l$ and
$\mathbft{X}^q$ are two $N\times L$ Toeplitz matrices corresponding
to the linear and quadratic parts defined as
\begin{align}
\mathbft{X}^l&:=\sqrt{\frac{3}{N}}\left[ \begin{array}{cccc}
x_0 & x_{-1} & \ldots & x_{-L+1}\\
x_1 & x_{0} & \ldots & x_{-L+2}\\
\vdots & \vdots &  & \vdots\\
x_{N-1} & x_{N-2} & \ldots & x_{N-L+1}
\end{array}\right]\label{eq:Xl_def}
\end{align}
\begin{align}
\mathbft{X}^q&:=\frac{3}{2}\sqrt{\frac{5}{N}}\left[ \begin{array}{cccc}
x_0^2-\frac{1}{3} & x_{-1}^2-\frac{1}{3} & \ldots & x_{-L+1}^2-\frac{1}{3}\\
x_1^2-\frac{1}{3} & x_{0}^2-\frac{1}{3} & \ldots & x_{-L+2}^2-\frac{1}{3}\\
\vdots & \vdots &  & \vdots\\
x_{N-1}^2-\frac{1}{3} & x_{N-2}^2-\frac{1}{3} & \ldots & x_{N-L+1}^2-\frac{1}{3}
\end{array}\right]\label{eq:Xq_def}
\end{align}
and $\mathbft{X}^b$ is a $N\times \frac{L(L-1)}{2}$ (non-Toeplitz)
matrix related to the bilinear part given by
\begin{align}
\mathbft{X}^b&:=\frac{3}{\sqrt{N}}\left[ \begin{array}{cccc}
x_0x_{-1} & x_{0}x_{-2} & \ldots & x_{-L+2}x_{-L+1}\\
x_1x_0 & x_{1}x_{-1} & \ldots & x_{-L+3}x_{-L+2}\\
\vdots & \vdots &  & \vdots\\
x_{N-1}x_{N-2} & x_{N-1}x_{N-3} & \ldots & x_{N-L+2}x_{N-L+1}
\end{array}\right].\label{eq:Xb_def}
\end{align}

Consider now the Grammian of $\mathbft{X}$, namely
$\mathbft{R}:=\mathbft{X}^T\mathbft{X}$. Comparing $\mathbf{X}$ with
$\mathbft{X}$, the columns of $\mathbft{X}$ have their $\ell_2$-norm
normalized in expectation, and thus $\mathbft{R}$ satisfies (p1).
Moreover, those columns of $\mathbft{X}$ corresponding to the
quadratic part (cf. submatrix $\mathbft{X}^q$) are shifted by the
variance of $x_n$. One can readily verify that (p2) is then
satisfied too.

The transition from $\mathbf{X}$ to $\mathbft{X}$ raises a
legitimate question though: Does the RIP of $\mathbft{X}$ provide
any insight on the compressed sampling guarantees for the original
Volterra problem? In the noiseless scenario, we actually substitute
the optimization problem in \eqref{eq:bp} by
\begin{equation}\label{eq:mbp}
\min_{\mathbft{h}}\left\{\|\mathbft{h}\|_1:~\mathbf{y}=\mathbft{X}\mathbft{h}\right\}.
\end{equation}
Upon matching the expansions
$\mathbf{X}\mathbf{h}=\mathbft{X}\mathbft{h}$, the following
one-to-one mapping holds
\begin{subequations}\label{eq:mapping}
\begin{align}
h_{0}&=\frac{1}{\sqrt{N}}\tilde{h}_0-\frac{1}{2}\sqrt{\frac{5}{N}}\sum_{k=1}^L\tilde{h}_2(k,k)\label{eq:mapping_c}\\
h_{1}(k)&=\sqrt{\frac{3}{N}}\tilde{h}_1(k),~k=1,\ldots,L\label{eq:mapping_l}\\
h_{2}(k,k)&=\frac{3}{2}\sqrt{\frac{5}{N}}\tilde{h}_2(k,k),~k=1,\ldots,L\label{eq:mapping_q}\\
h_{2}(k_1,k_2)&=\frac{3}{\sqrt{N}}\tilde{h}_2(k_1,k_2),~k_1=1,\ldots,L,~k_2=k_1+1,\ldots,L.\label{eq:mapping_b}
\end{align}
\end{subequations}
It is now apparent that a sparse solution of \eqref{eq:mbp}
translates to a sparse solution of \eqref{eq:bp} except for the
constant term in \eqref{eq:mapping_c}. By deterministically
adjusting the weights $\{w_i\}_{i=1}^{M}$ and the parameter
$\lambda_N$ in \eqref{eq:Lasso}, this argument carries over to the
Lasso optimization problem and answers affirmatively the previously
posed question. Note though that such a modification serves only
analytical purposes; practically, there is no need to solve the
modified compressed sampling problems.

\begin{remark}\label{re:wiener}
Interestingly, transition from the original Volterra matrix to the
modified one resembles the replacement of the Volterra by the Wiener
polynomials for nonlinear system identification \cite{MaSi00}.
Wiener polynomials are known to facilitate mean-square error
(MSE)-optimal estimation of Volterra modules for a white Gaussian
input; see e.g., \cite{MaSi00}. Our modification, adjusted to a
uniformly distributed input, facilitates the RIP analysis of the
Volterra regression matrix.
\end{remark}

One of the main results of this paper is summarized in the following
theorem (see the Appendix for a proof).

\begin{theorem}[RIP in Volterra Filtering]\label{th:rip_volterra}
Let $\{x_i\}_{i=-L+1}^N$ be an input sequence of independent
random variables drawn from $\mathcal{U}[-1,1]$, and define
$M:=(L+1)(L+2)/2$. Assume that the $N\times M$ modified
Volterra regression matrix $\mathbft{X}$ defined in
\eqref{eq:Xt_def}-\eqref{eq:Xb_def} is formed by such an
input for $L\geq 7$ and $N\geq 160$. Then, for any
$\delta_s\in(0,1)$ and for any $\gamma\in(0,1)$, whenever
$N\geq \frac{5C}{(1-\gamma)\delta_s^2}\cdot s^2 \log L$,
the matrix $\mathbft{X}$ possesses RIP $\delta_s$ for
$s\geq 2$ with probability exceeding
$1-\exp\left(-\frac{\gamma\delta_s^2}{C}\cdot \frac{N}{s^2}\right)$,
where $C=2,835$.
\end{theorem}

The theorem asserts that an order $s^2\log L$ observations suffice
to recover an $s$-sparse non-homo\-gene\-ous second-order Volterra
filter of memory $L$ probed by a uniformly distributed input scales
as $s^2\log L$. Since the number of unknowns $M$ is
$\mathcal{O}(L^2)$, the bound on $N$ scales also as $s^2\log M$. The
bound agrees with the bounds obtained for the linear filtering setup
\cite{HaNo10}, whereas now the constants are larger due to the more
involved dependencies among the entries of the associated regression
matrix.

\subsection{RIP for Multivariate Polynomial Regression}\label{subsec:rip_poly}
Consider now the case where $f(\mathbf{x})$ describes a sparse
linear-quadratic model
\begin{equation}\label{eq:lq_reg1}
f(\mathbf{x}_1)=h_0 + \sum_{k=1}^L h_1(k) x_k + \sum_{k_1=1}^L\sum_{k_2=k_1}^L h_2(k_1,k_2)x_{k_1}x_{k_2}.
\end{equation}
Given $N$ output samples $\{y(n)\}_{n=1}^N$, corresponding to input
data $\{\mathbf{x}_1(n)\}_{n=1}^N$ drawn independently from
$\mathcal{U}\left[-1,1\right]^L$, the goal is to recover the sparse
$M\times 1$ vector $\mathbf{h}$ comprising the $h_1(k)$'s and
$h_2(k_1,k_2)$'s. Note that $M=(L+1)(L+2)/2$ here. As explained in
Section \ref{sec:problem}, the noiseless expansion in
\eqref{eq:lq_reg1} can be written as
$\mathbf{y}=\mathbf{X}\mathbf{h}$; but, contrary to the Volterra
filtering setup, the rows of $\mathbf{X}$ are now statistically
independent. The last observation differentiates significantly the
RIP analysis for polynomial regression and leads to tighter
probabilistic bounds.

Our analysis builds on \cite{Ra10}, which deals with finding a
sparse expansion of a function $f(\mathbf{x})=\sum_{t=1}^T c_t
\psi_t(\mathbf{x})$ over a bounded orthonormal set of functions
$\{\psi_t(\mathbf{x})\}$. Considering $\mathcal{D}$ a measurable
space, e.g., a measurable subset of $\mathbb{R}^L$ endowed with a
probability measure $\nu$, the set of functions
$\{\psi_t(\mathbf{x}):\mathcal{D}\rightarrow \mathbb{R}\}_{t=1}^T$
is a bounded orthonormal system if for all $t_1,t_2=1,\ldots,T$
\begin{equation}\label{eq:bo_def1}
\int_{\mathcal{D}} \psi_{t_1}(\mathbf{x})\psi_{t_2}(\mathbf{x})d\nu(\mathbf{x})=\delta_{t_1,t_2}
\end{equation}
where $\delta_{t_1,t_2}$ denotes the Kronecker delta function, and
for some constant $K\geq 1$ it holds that
\begin{equation}\label{eq:bo_def2}
\sup_{t} \sup_{\mathbf{x}\in \mathcal{D}} |\psi_t(\mathbf{x})|\leq K.
\end{equation}
After sampling $f(\mathbf{x})$ at $\{\mathbf{x}(n)\in
\mathcal{D}\}_{n=1}^N$, the involved $N\times T$  regression matrix
$\mathbf{\Psi}$ with entries $\Psi_{n,t}:= \psi_t\left(\mathbf{x}(n)
\right)$ admits the following RIP characterization \cite[Theorems
4.4 and 8.4]{Ra10}.
\begin{theorem}[RIP in bounded orthonormal systems \cite{Ra10}] \label{th:rip_bounded}
Let $\mathbf{\Psi}$ be the $N\times T$ matrix associated with a bounded orthonormal system with constant $K\geq 1$ in \eqref{eq:bo_def2}. Then, for any $\delta_s\in\left(0,0.5\right]$, there exist universal positive constants $C$ and $\gamma$, such that whenever $N\geq \frac{CK^2}{\delta_s^2}\cdot s\log^4T$, the matrix $\frac{1}{\sqrt{N}}\mathbf{\Psi}$ possesses RIP $\delta_{s}$ with probability exceeding $1-\exp\left(-\frac{\gamma\delta_s^2}{CK^2}\cdot\frac{N}{s}\right)$.
\end{theorem}

In the linear-quadratic regression of \eqref{eq:lq_reg1}, even
though the basis functions $\left\{1,\{x_i\}, \{x_{i_1}x_{i_2}\}
\right\}$ are bounded in $[-1,1]^L$, they are not orthonormal in the
uniform probability measure. Fortunately, our input transformation
trick devised for the Volterra filtering problem applies to the
polynomial regression too. The expansion is now over the basis
functions $\{\psi_m(\mathbf{x})\}_{m=1}^M$
\begin{equation}\label{eq:lq_basis}
\left\{1,   \{\sqrt{3}x_i\}, \left\{\frac{3\sqrt{5}}{2}\left(x_i^2-\frac{1}{3}\right)\right\}, \{3x_{i_1}x_{i_2}\} \right\}
\end{equation}
where the last subset contains all the unique, two-variable
monomials lexicographically ordered. Upon stacking the function
values $\{y_n\}_{n=1}^N$ in $\mathbf{y}$ and properly defining
$\mathbft{h}$, the expansion $\mathbf{y}=\mathbf{X}\mathbf{h}$ can
be replaced by $\mathbf{y}=\mathbft{X}\mathbft{h}$, where the
entries of $\mathbft{X}$ are
\begin{equation}\label{eq:lq_regX}
\tilde{X}_{n,m}:=\frac{\psi_m\left(\mathbf{x}(n)\right)}{\sqrt{N}}.
\end{equation}
Vectors $\mathbf{h}$ and $\mathbft{h}$ are related through the
one-to-one mapping in \eqref{eq:mapping}; thus, sparsity in one is
directly translated to the other. Identifiability of a sparse
$\mathbf{h}$ can be guaranteed by the RIP analysis of $\mathbft{X}$
presented in the next lemma.

\begin{lemma}[RIP in linear-quadratic regression]\label{le:rip_lq}
Let $x_i(n)$ for $i=1,\ldots,L$ and $n=1,\ldots,N$ independent
random variables uniformly distributed in $[-1,1]$, and define
$M:=(L+1)(L+2)/2$. Assume that the $N\times M$ modified polynomial
regression matrix $\mathbft{X}$ in \eqref{eq:lq_regX} is generated
by this sequence for $L\geq 4$. Then, for any $\delta_s\in(0,0.5]$,
there exist universal positive constants $C$ and $\gamma$, such that
whenever $N\geq \frac{144C}{\delta_s^2}\cdot s\log^4L$, the matrix
$\mathbft{X}$ possesses RIP $\delta_{s}$ with probability exceeding
$1-\exp\left(-\frac{\gamma\delta_s^2}{9C}\cdot\frac{N}{s}\right)$.
\end{lemma}

\begin{IEEEproof}
The inputs $\mathbf{x}(n)$ are uniformly drawn over
$\mathcal{D}=[-1,1]^L$, and it is easy to verify that the basis
functions $\{\psi_m(\mathbf{x})\}_{m=1}^{M}$ in \eqref{eq:lq_basis}
form a bounded orthonormal system with $K=3$. Hence, Theorem
\ref{th:rip_bounded} can be straightforwardly applied. Since $M\leq
L^2$ for $L\geq 4$, it follows that $\log^4 M<16\log^4 L$.
\end{IEEEproof}

Lemma \ref{le:rip_lq} assures that an $s$-sparse linear-quadratic
$L$-variate expansion with independent uniformly distributed inputs
can be identified with high probability from a minimum number of
observations that scales as $s\log^4 L$ or $s\log^4 M$. Comparing
this to Theorem \ref{th:rip_volterra}, the bound here scales
linearly with $s$. Moreover, except for the increase in the power of
the logarithmic factor, the bound is close to the one obtained for
random Gaussian and Bernoulli matrices. The improvement over the
Volterra RIP bound is explained by the simpler structural dependence
of the matrix $\mathbf{X}$ involved.

Another interesting polynomial regression paradigm is when the
nonlinear function $f(\mathbf{x}_1)$ admits a sparse polynomial
expansion involving $L$ inputs, and all products up to $P$ of these
inputs, that is
\begin{align}\label{eq:gene_expansion}
f(\mathbf{x}_1)&=h_0 + \sum_{k=1}^L h_1(k) x_k + \sum_{k_1=1}^L\sum_{k_2=k_1+1}^L h_2(k_1,k_2)x_{k_1}x_{k_2}+\ldots\\
&+\sum_{k_1=1}^L\sum_{k_2=k_1+1}^L\ldots\sum_{k_P=k_{P-1}+1}^L h_P(k_1,k_2,\cdots,k_P)x_{k_1}x_{k_2}\ldots x_{k_P}.\nonumber
\end{align}
This is the typical multilinear regression setup appearing in GWA studies \cite{WuCh09}, \cite{Co09}. Because there are $\binom{L}{p}$ monomials of order $p$, the vector $\mathbf{h}$ comprising all the expansion coefficients has dimension
\begin{equation}\label{eq:gene_dimension}
M=\sum_{p=0}^P\dbinom{L}{p}\leq (L+1)^P
\end{equation}
where the last inequality provides a rough upper bound. The goal is
again to recover an $s$-sparse $\mathbf{h}$ given the sample
phenotypes $\{y_n\}_{n=1}^N$ over the genotype values
$\{\mathbf{x}_1(n)\}_{n=1}^N$. Vectors $\mathbf{x}_1(n)$ are drawn
either from $\{-1,0,1\}^L$ or $\{-1,1\}^L$ depending on the assumed
genotype model (additive for the first alphabet; and dominant or
recessive for the latter) \cite{WuCh09}. Without loss of generality,
consider the ternary alphabet with equal probabilities. Further,
suppose for analytical convenience that the entries of
$\mathbf{x}_1(n)$ are independent. Note that the input has mean zero
and variance $2/3$.

The RIP analysis for the model in \eqref{eq:gene_expansion} exploits
again Theorem \ref{th:rip_bounded}. Since now every single input
appears only linearly in \eqref{eq:gene_expansion}, the basis
functions $\left\{1,\{x_i\}, \{x_{i_1}x_{i_2}\},\ldots\right\}$ are
orthogonal w.r.t. the assumed point mass function. A bounded
orthonormal system $\{\psi_m\left({\mathbf{x}}\right)\}_{m=1}^M$ can
be constructed after scaling as
\begin{equation}\label{eq:gene_basis}
\left\{1, \{(2/3)^{-1/2}x_{i_1}\}, \{(2/3)^{-2/2}x_{i_1}x_{i_2}\},\ldots, \{(2/3)^{-P/2}x_{i_1}x_{i_2}\cdots x_{i_P}\}  \right\}
\end{equation}
while the set is bounded by $K=(3/2)^{P/2}$. Similar to the
linear-quadratic case in \eqref{eq:lq_reg1}, the original
multilinear expansion $\mathbf{X}\mathbf{h}$ is transformed to
$\mathbft{X}\mathbft{h}$, where $\mathbft{X}$ is defined as in
\eqref{eq:lq_regX} with the new basis of \eqref{eq:gene_basis}, and
$\mathbft{h}$ is an entry-wise rescaled version of $\mathbf{h}$.
Based on these facts, the RIP characterization of $\mathbft{X}$
follows readily from the ensuing lemma.\footnote{After our
conference precursor \cite{KeAnGia09}, we became aware of a recent
result in \cite{NaNo10}, which relates to Lemma \ref{le:rip_ml}. The
differences are: i) only the $P$-th order term in expansion
\eqref{eq:gene_expansion} is considered in \cite{NaNo10}; and ii)
inputs $\{x_i(n)\}$ adhere to the binary $\{\pm 1\}$ alphabet in
\cite{NaNo10}, as opposed to the ternary one in Lemma
\ref{le:rip_ml}.}

\begin{lemma}[RIP in multilinear expansion]\label{le:rip_ml} Let $x_i(n)$ for $i=1,\ldots,L$ and $n=1,\ldots,N$ independent random variables equiprobably drawn from $\{-1,0,1\}$, and $M$ defined as in \eqref{eq:gene_dimension}. The $N\times M$ modified multilinear regression matrix $\mathbft{X}$ in \eqref{eq:lq_regX} and \eqref{eq:gene_basis} is generated by this sequence. Then, for any $\delta_s\in(0,0.5]$, there exist universal positive constants $C$ and $\gamma$, such that whenever $N\geq \frac{C}{\delta_s^2} \left(\frac{3}{2}\right)^P P^4 s\log^4(L+1)$, the matrix $\mathbft{X}$ possesses RIP $\delta_s$ with probability exceeding $1-\exp\left(-\frac{\gamma\delta_s^2}{C (3/2)^P}\cdot\frac{N}{s}\right)$.
\end{lemma}

Since $P$ is often chosen in the order of 2 due to computational
limitations, Lemma \ref{le:rip_ml} guarantees the RIP to hold with
high probability when the number of phenotype samples $N$ scales at
least as $s\log^4 L$.

\section{Simulated Tests}\label{sec:simulations}
The RIP analysis performed in the previous section provides
probabilistic bounds on the identifiability of sparse polynomial
representations. In this section, we evaluate the applicability of
sparsity-aware polynomial estimators using synthetic and real data.
The experimental results indicate that sparsity-promoting recovery
methods attain accurate results even when the number of measurements
is less than the RIP-derived bounds, and, in any case, they
outperform the sparsity-agnostic estimators.

\subsection{Batch and Adaptive Volterra Filters}
We first focus on the sparse Volterra system identification setup.
The system under study was an LNL one, consisting of a linear filter
with impulse response $\mathbf{h}_f=\left[ 0.36 ~ 0 ~ 0.91 ~ 0 ~ 0 ~
0.19\right]^T$, in cascade with the memoryless nonlinearity
$f(x)=-0.5x^3+0.4x^2+x$, and the same linear filter. This system is
exactly described by a Volterra expansion with $L=11$ and $P=3$,
leading to a total of $M=\binom{L+P}{P}=364$ coefficients collected
in the vector $\mathbf{h}_0$. Out of the $364$ coefficients only
$48$ are nonzero. The system input was modeled as $x(n)\sim
\mathcal{N}(0,1)$, while the output was corrupted by additive noise
$v(n)\sim \mathcal{N}(0,0.1)$. First, the batch estimators of
Section \ref{subsec:batch} were tested, followed by their sequential
counterparts.

\begin{figure}
\centering
\subfigure{
\includegraphics[width=0.45\linewidth]{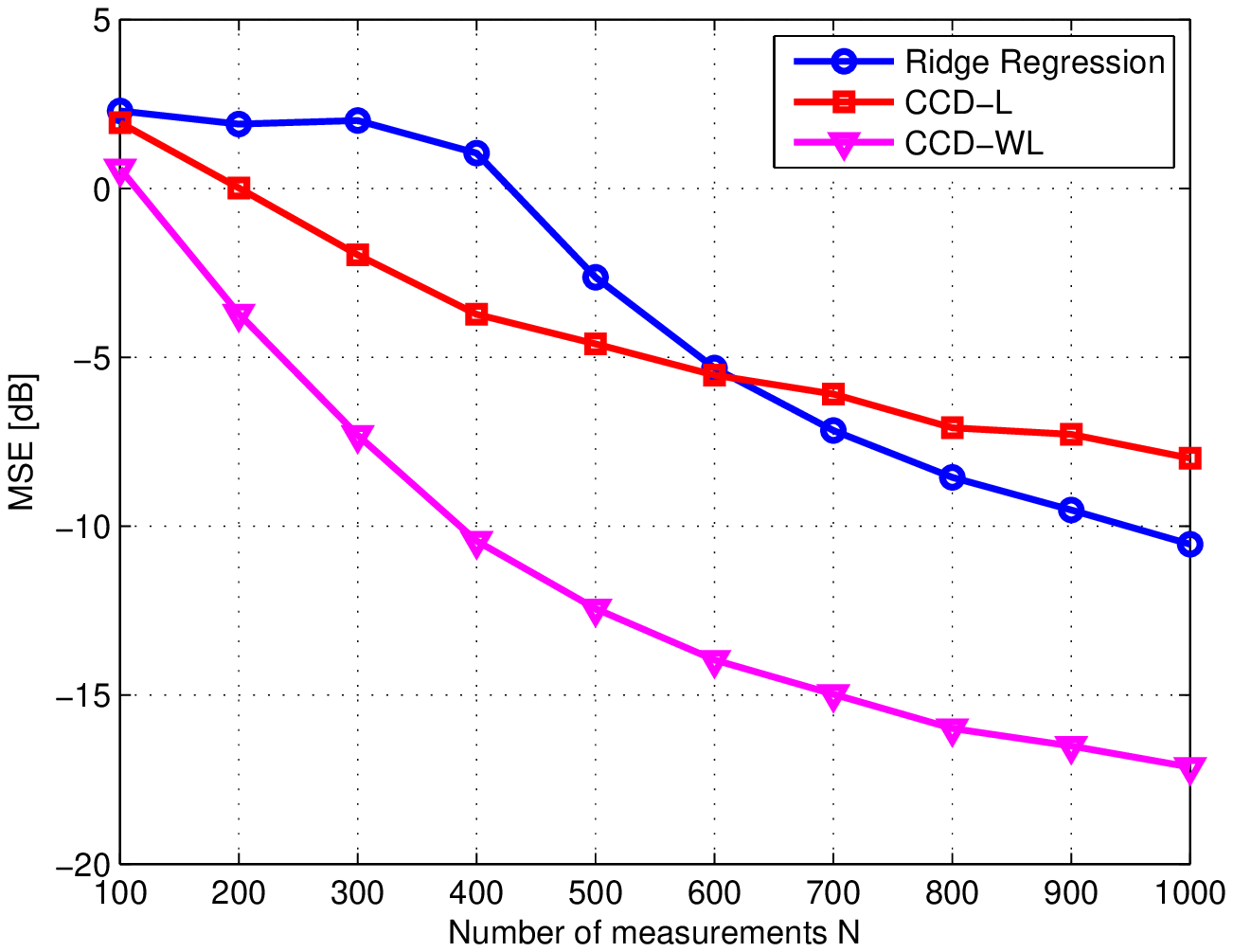}
\label{subfig:batch}}
\subfigure{
\includegraphics[width=0.45\linewidth]{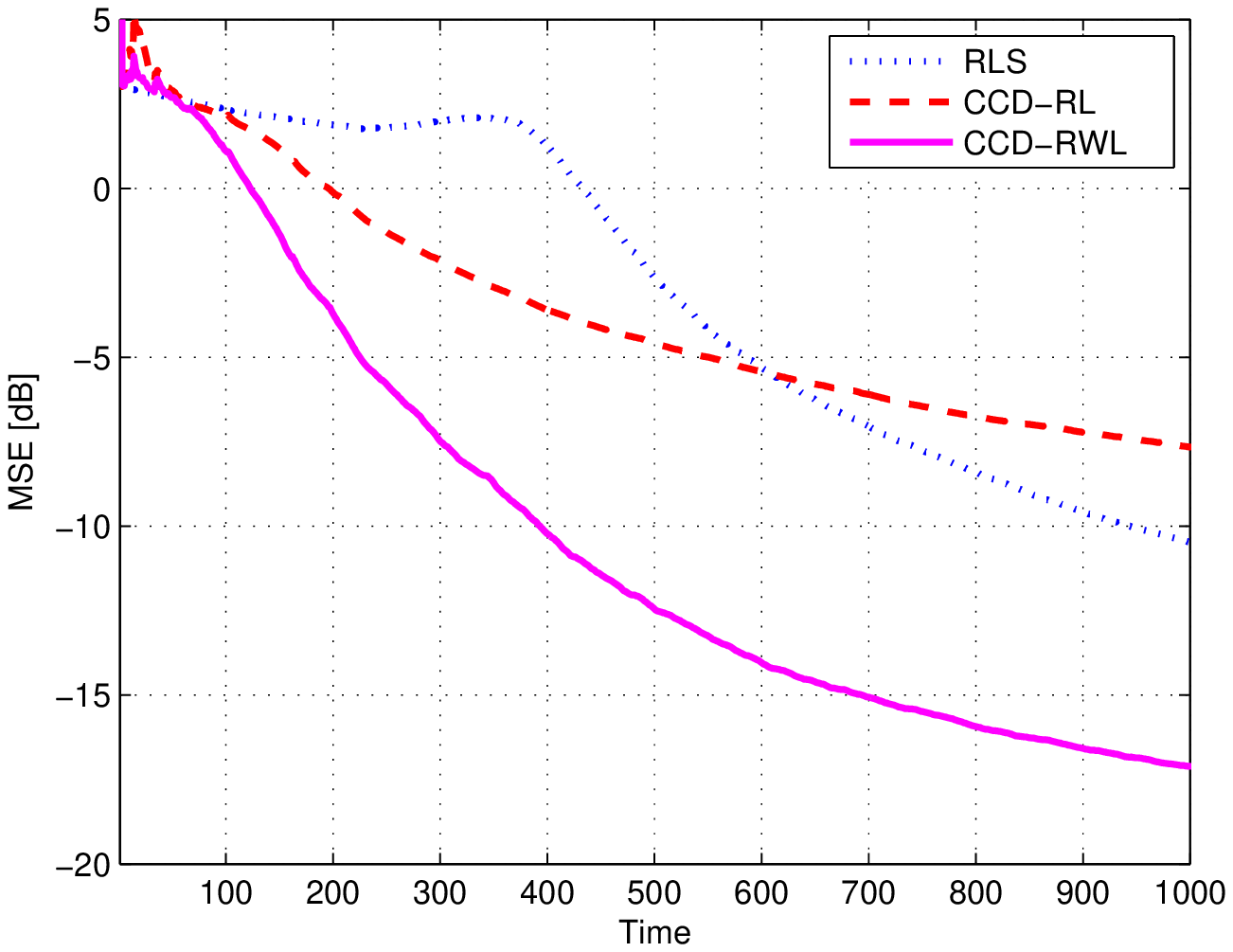}
\label{subfig:adaptive}}
\caption{MSE of (a) batch and (b) adaptive Volterra estimators.}
\label{fig:estimators}
\end{figure}

In Fig.~\ref{subfig:batch}, the obtained MSE,
$\mathbb{E}\left[\|\mathbf{h}_0- \hat{\mathbf{h}}\|_2^2\right]$,
averaged over $100$ Monte Carlo runs, is plotted against the number
of observations, $N$, for the following estimators: (i) the ridge
estimator of \eqref{eq:ridge} with $\delta{=}1$; (ii) the Lasso
(CCD-L) estimator with $\lambda_N{=}0.7\sqrt{N}$; and, (iii) the
weighted Lasso (CCD-WL) estimator with $\lambda_N{=}0.08\log N$. The
scaling rules for the two $\lambda_N$s follow the results of
\cite{AnBaGia10} and \cite{Zou06}. It can be seen that the
sparsity-agnostic ridge estimator is outperformed by the Lasso
estimator for short observation intervals ($N{<}600$). For larger
$N$, where $\mathbf{X}^T\mathbf{X}$ becomes well-conditioned, the
former provides improved estimation accuracy. However, CCD-WL offers
the lowest MSE for every $N$, and provides reasonably accurate
estimates even for the under-determined case $(N{<}364)$.


Performance of the sequential estimator in Section
\ref{subsec:sequential} was assessed in the same setup.
Fig.~\ref{subfig:adaptive} illustrates the MSE convergence, averaged
over 100 Monte Carlo runs, for the following three recursive
algorithms: (i) the conventional RLS of \eqref{eq:RLS-opt}; (ii) the
cyclic coordinate descent recursive Lasso (CCD-RL); and, (iii) its
weighted version (CCD-RWL). Since the system was time-invariant, the
forgetting factor was set to $\beta=1$. It can be observed that the
conclusions drawn for the batch case carry over to the recursive
algorithms too. Moreover, a comparison of Figs.~\ref{subfig:batch}
and \ref{subfig:adaptive} indicates that the sparsity-aware iterates
of Table \ref{alg:recursive} approximate closely the exact per time
instance problem in \eqref{eq:RLasso}.


\subsection{Multilinear Regression for GWA Analysis}
Here we test sparse polynomial modeling for studying the epistatic
effects in quantitative trait analysis. In quantitative genetics,
the phenotype is a quantitative trait of an organism, e.g., the
weight or height of barley seeds \cite{TiMa96}. Ignoring
environmental effects, the phenotype is assumed to follow a linear
regression model over the individual's genotype, including
single-gene (main) and gene-gene (epistatic) effects \cite{Xu07},
\cite{Co09}. The genotype consists of markers which are samples of
chromosomes taking usually binary $\{\pm 1\}$ values. Determining
the so-called quantitative trait loci (QTL) corresponds to detecting
the genes and pairs of genes associated with a particular trait
\cite{Xu07}. Since the studied population $N$ is much smaller than
the number of regressors $M$, and postulating that only a few
genotype effects determine the trait considered, QTL analysis falls
under the sparse multilinear (for $P=2$) model of
\eqref{eq:gene_expansion}.

\subsubsection{Synthetic Data}
The first QTL paradigm is a synthetic study detailed in \cite{Xu07}.
A population of $N$=600 individuals is simulated for a chromosome of
1800 cM (centiMorgan) evenly sampled every 15 cM to yield $L$ = 121
markers. The true population mean and variance are 5.0 and 10.0,
respectively. The phenotype is assumed to be linearly expressed over
the intercept, the $L$ main effects, and the $\binom{L}{2}=7,260$
epistatic effects, leading to a total of $M=7,382$ regressors. The
QTLs simulated are 9 single markers and 13 marker pairs. Note that
the simulation accommodates markers (i) with main only, (ii)
epistatic only, and (iii) both main and epistatic effects. Since the
intercept is not regularized, genotype and phenotype data were
centered, i.e., their sample mean was subtracted, and the intercept
was determined at the end as the sample mean of the initial I/O data
on the fitted model.


Parameters $\delta$ and $\lambda$ for ridge and (w)Lasso estimators,
respectively, were tuned through 10-fold cross-validation over an
100-point grid \cite{HaTiFr09}; see Table \ref{tbl:synthetic}. The
figure of merit for selecting the parameters was the prediction
error (PE) over the unseen data, i.e., $\sum_{v=1}^{V}\|
\mathbf{y}_v - \mathbf{X}_v \hat{\mathbf{h}}_v \|_2^2 / (N/V)$,
where $V=10$ and $\hat{\mathbf{h}}_v$ is the regression vector
estimated given all but the $(\mathbf{y}_v, \mathbf{X}_v)$
validation data. The value of $\delta$ attaining the smallest PE was
subsequently used for determining the weights for the wLasso
estimator. Having tuned the regularization parameters, the MSE
provided by the three methods was averaged over 100 Monte Carlo runs
on different phenotypic data while keeping the genotypes fixed. The
(w)Lasso estimators were run using the glmnet software
\cite{FrHaHoTi07}. Each of the three algorithms took less than 1 min
and 1 sec for cross-validation and final estimation, respectively.

\begin{figure}
\centering
\subfigure[True model]{
\includegraphics[width=0.40\linewidth]{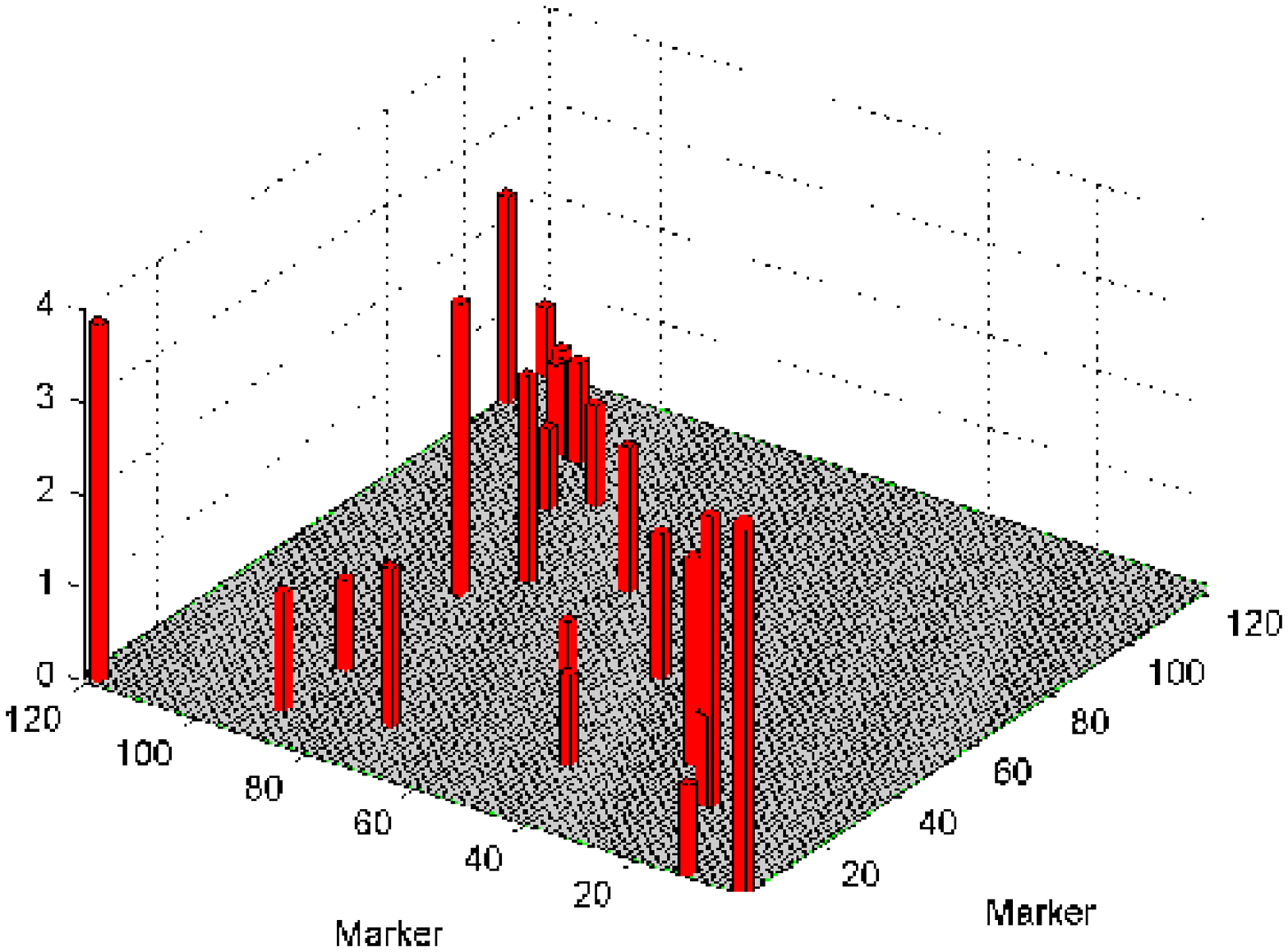}
\label{subfig:sTrue}}
\subfigure[Ridge regression]{
\includegraphics[width=0.40\linewidth]{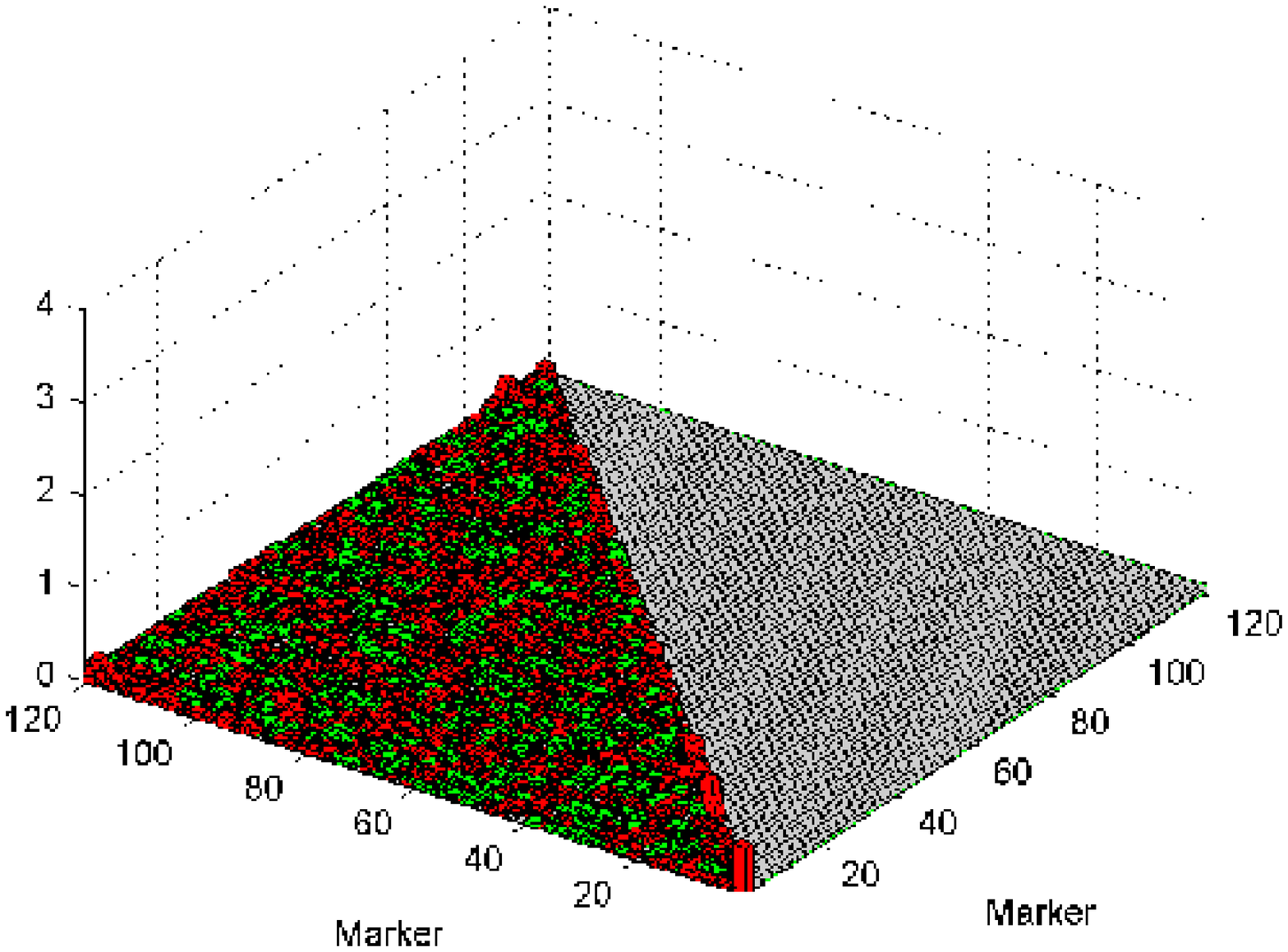}
\label{subfig:sRidge}}\\
\subfigure[Lasso]{
\includegraphics[width=0.40\linewidth]{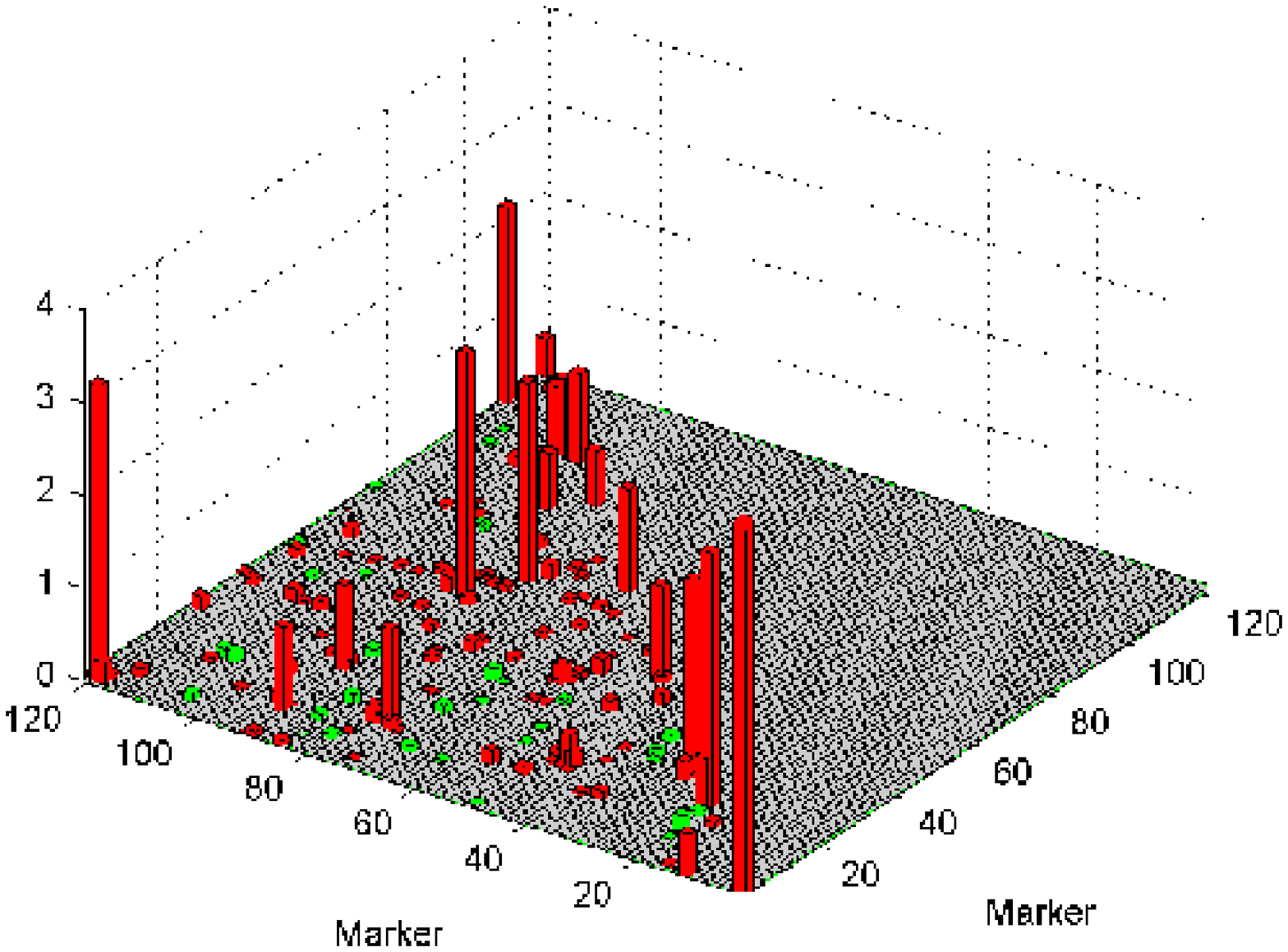}
\label{subfig:sLasso}}
\subfigure[wLasso]{
\includegraphics[width=0.40\linewidth]{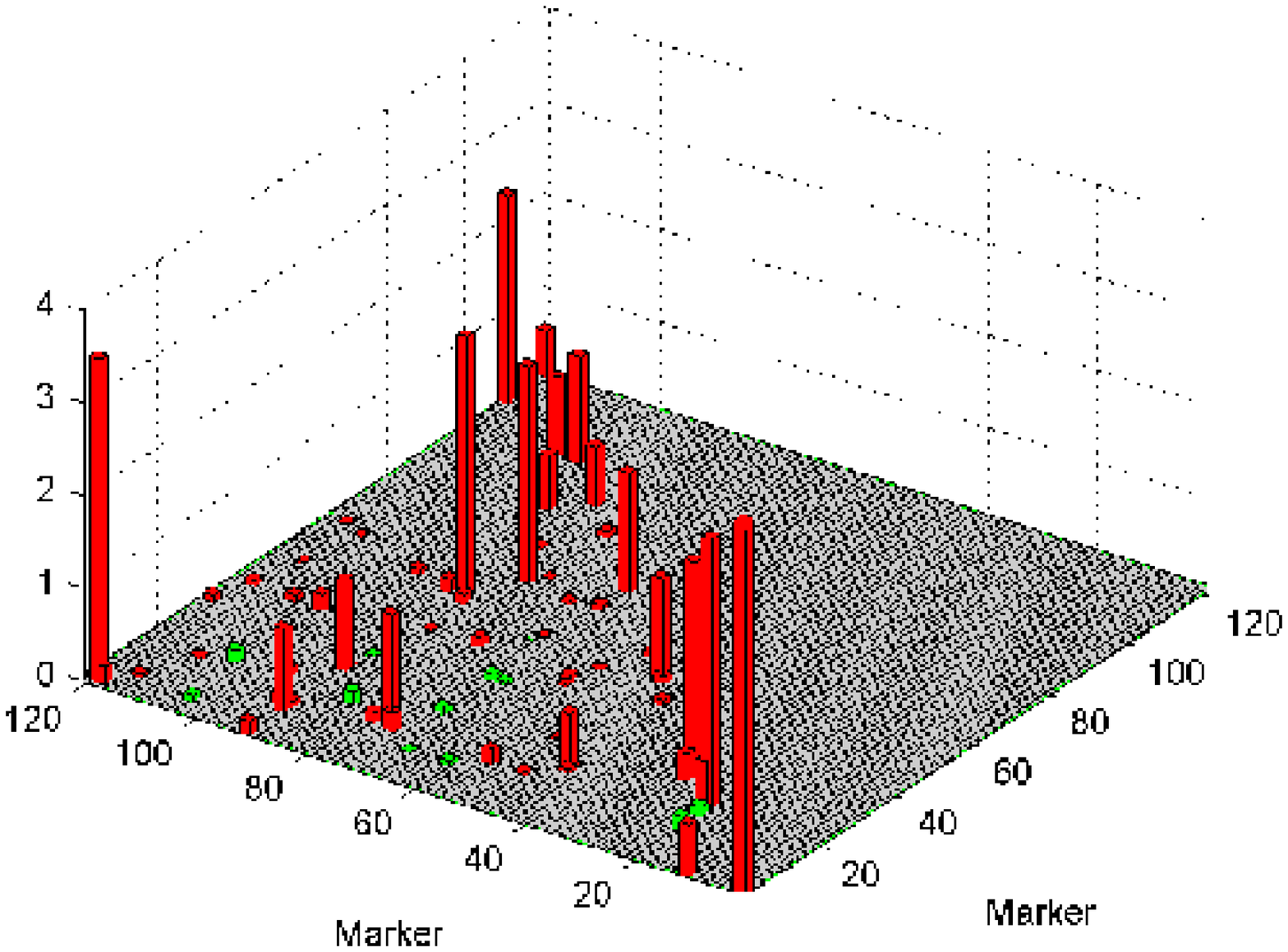}
\label{subfig:swLasso}} \caption{Regression vector estimates for the
synthetic gene data. The main (epistatic) effects are shown on the
diagonal (left diagonal part), while red (green) bars correspond to
positive (negative) entries.} \label{fig:synthetic}
\end{figure}

As can be seen from Table \ref{tbl:synthetic}, Lasso attains the
smaller PE. However, wLasso provides significantly higher estimation
accuracy at a PE value comparable to Lasso. The number of non-zero
regression coefficients indicated in the fourth column shows that
ridge regression yields an over-saturated model. As shown more
clearly in Fig.~\ref{fig:synthetic}, where the true and the
estimated models are plotted, the wLasso yields a sparser, closer to
the true model, while avoiding some spurious coefficients found by
Lasso.


\subsubsection{Real data from a barley experiment}
The second QTL experiment entails a real dataset collected by the
North American Barley Genome Mapping Project as described in
\cite{TiMa96}, \cite{XuJi07}, and outlined shortly next. Aiming at a
GWA analysis on barley height (HGT), the population consists of
$N$=145 doubled-haploid lines of a cross between two barley lines,
Harrington and TR306. The height of each individual was measured
under 27 different environments, and the phenotype was taken to be
the sample average. There are $L$ = 127 markers covering a 1270 cM
segment of the genome with an average marker interval of 10.5 cM.
The genotype is binary: +1 (-1) for the TR306 (Harrington) allele.
There is a $5\%$ of missing values which are modeled as zeros in
order to minimize their effect \cite{Xu07}. The main and epistatic
QTL analysis involves $M=1+127+\binom{127}{2}=8,129$ regressors.

\begin{figure}
\centering
\subfigure[Lasso]{
\includegraphics[width=0.40\linewidth]{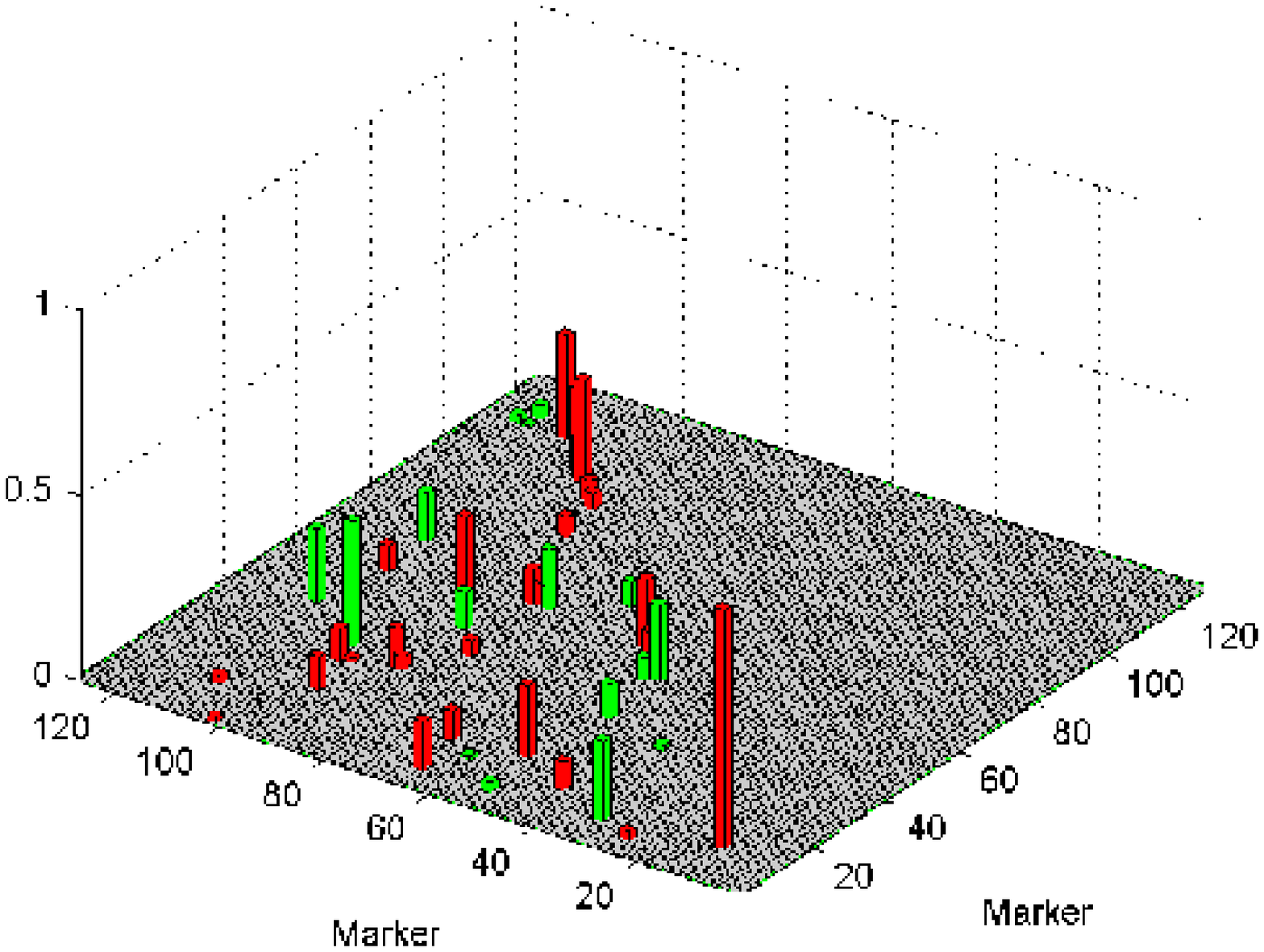}
\label{subfig:rLasso}}
\subfigure[wLasso]{
\includegraphics[width=0.40\linewidth]{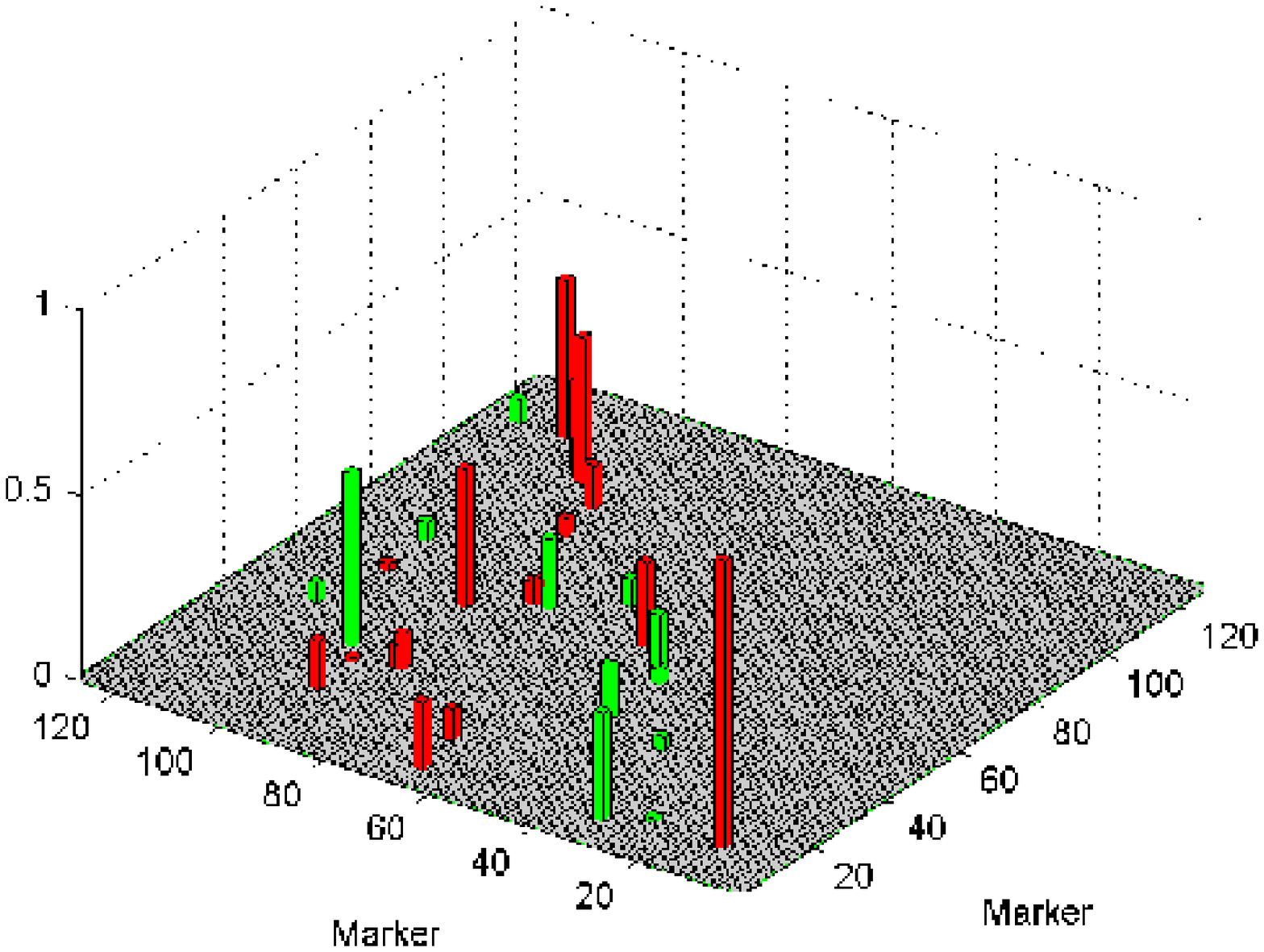}
\label{subfig:rwLasso}} \caption{Regression vector estimates for the
real QTL barley data. The main (epistatic) effects are shown on the
diagonal (left diagonal part), while red (green) bars correspond to
positive (negative) entries.} \label{fig:real}
\end{figure}

The regularization parameter values were selected through
leave-one-out cross-validation \cite{HaTiFr09}; see Table
\ref{tbl:real}. The ridge estimator fails to handle over-fitting and
$\delta$ is set to a large value yielding regression coefficients of
insignificant amplitude. Using the ridge estimates to weight the
regression coefficients, wLasso yields a PE slighty smaller than the
one attained by Lasso; but it reduces the spurious coefficients. As
shown in Fig.~\ref{fig:real}, wLasso provides a more parsimonious
model with fewer spurious peaks than the Lasso-inferred model.
Closer investigation of the wLasso QTLs exceeding $0.1$ in
magnitude, shown in Table \ref{tbl:real2}, offers the following
interesting observations: (i) epistatic effects are not negligible;
(ii) there are epistatic effects related to QTLs with main effects,
e.g., the $(35,99)$ pair is related to marker $(101)$; (iii) there
are epistatic effects such as the $(9,33)$ one involving markers
with no main effect.

\section{Conclusions}\label{sec:conclusions}
The idea of exploiting sparsity in the representation of a system,
already widely adopted for linear regression and system
identification, has been permeated here to estimate sparse Volterra
and polynomial models. The abundance of applications allowing for an
interpretative parsimonious polynomial expansion and the inability
of kernel regression to yield such an expansion necessitate
sparsity-aware polynomial estimators. This need was successfully met
here both from practical and analytical perspectives.
Algorithmically, the problem was solved via the batch (weighted)
Lasso estimators, where for the weighted one, the weights were
efficiently found through the kernel trick. To further reduce the
computational and memory load and enable tracking, an adaptive
sparse RLS-type algorithm was devised. On the analytical side, RIP
analysis was carried out for the two models. It was shown that an
$s$-sparse linear-quadratic Volterra filter can be recovered with
high probability using measurements in the order of $s^2\log L$; a
bound that interestingly generalizes the results from the linear
filtering problem to the Volterra one. For the sparse polynomial
expansions considered, the bound improved to $s\log^4 L$, which also
generalizes the corresponding linear regression results. The
potential of the aforementioned sparse estimation methods was
numerically verified through synthetic and real data. The developed
sparse adaptive algorithms converged fast to the exact solution,
while the (weighted) Lasso estimators outperformed the LS-based one
in all simulated scenarios, as well as in the GWA study on real
barley data. Future research directions include extending the bounds
derived to higher-order models, and utilizing our adaptive methods
to accomplish epistatic GWA studies on the considerably higher
dimensional human genome.

\appendix
Outlining some tools regarding concentration inequalities precede
the proof of Theorem \ref{th:rip_volterra}.


\begin{lemma}[Hoeffding's inequality]\label{le:hoeff}
Given $t>0$ and independent random variables $\{x_i\}_{i=1}^N$
bounded as $a_i\leq x_i \leq b_i$ almost surely, the sum
$s_N:=\sum_{i=1}^N x_i$ satisfies
\begin{equation}\label{eq:hoeff}
\pr\left(|s_N-\mathbb{E}[s_N]|\geq t\right)\leq 2 \exp \left( -\frac{2t^2}{\sum_{i=1}^N(b_i-a_i)^2} \right).
\end{equation}
\end{lemma}

It is essentially a Chernoff-type result on the concentration of a
sum of independent bounded random variables around its mean.
However, the subsequent analysis on the RIP of the Volterra filter
considers sums of structurally dependent random variables. Useful
probability bounds on such sums can be derived based on the
following lemma.

\begin{lemma}[Hoeffding's inequality with dependent summands \cite{Pe01}]\label{le:hoeff_d} Consider random variables $\{x_i\}_{i=1}^N$ bounded as $a\leq x_i \leq b$ almost surely. Assume also they can be partitioned into $M$ collectively exhaustive and mutually exclusive subsets $\{\mathcal{N}_m\}_{m=1}^M$ with respective cardinalities $\{N_m\}_{m=1}^M$ such that the variables within each subset are independent. Then, for any $t>0$ the sum $s_N:=\sum_{i=1}^N x_i$ satisfies
\begin{equation}\label{eq:hoeff_d}
\pr\left(|s_N-\mathbb{E}[s_N]|\geq t\right)\leq 2M \exp \left( -\frac{2t^2}{N^2(b-a)^2}N_{\min} \right)
\end{equation}
where $N_{\min}:=\min_{m}\{N_m\}$.
\end{lemma}

Note that the sharpness of the bound in \eqref{eq:hoeff_d} depends
on the number of subsets $M$ as well as the minimum of their
cardinalities $N_{\min}$. One should not only strive for the minimum
number of intra-independent subsets, but also arrange $N_m$'s as
uniformly as possible. For example, partitioning with the minimum
number of subsets may yield $N_{\min}=1$ that corresponds to a loose
bound.

The partitioning required in Lemma \ref{le:hoeff_d} is not always
easy to construct. An interesting way to handle this construction is
offered by graph theory as suggested in \cite{Pe01}. The link
between structural dependencies in a set of random variables
$\{x_i\}_{i=1}^N$ and graph theory hinges on their
\textit{dependency graph} $G$. The latter is defined as the graph
having one vertex per $x_i$, and an edge between every pair of
vertices corresponding to dependent $x_i$'s. Recall that the degree
of a vertex is the number of edges attached to it, and the degree of
a graph $\Delta(G)$ is the maximum of the vertex degrees. Finding
group-wise statistical independence among random variables can be
seen as a coloring of the dependency graph. The problem of coloring
aims at assigning every vertex of a graph to a color (class) such
that there are no adjacent vertices sharing the same color.
Moreover, coloring of a graph is equitable if the cardinality of
every color does not differ by more than one from the cardinalities
of every other color. Thus, an $M$-equitable coloring of the
dependency graph means that the random variables can be partitioned
in $M$ intra-independent subsets whose cardinalities are either
$\left\lfloor \frac{N}{M}\right\rfloor$ or $\left\lfloor
\frac{N}{M}\right\rfloor+1$. A key theorem by Hajnal and Szemeredi
guarantees that a graph $G$ has an $M$-equitable coloring for all
$M\geq \Delta(G)+1$; see e.g., \cite{Pe01}. Combining this result
with Lemma \ref{le:hoeff_d}, yields the following corollary.

\begin{corollary}[Hoeffding's inequality and dependency graph \cite{Pe01}, \cite{HaNo10}]\label{co:hoeff_dd} Consider random variables $\{x_i\}_{i=1}^N$ bounded as $a\leq x_i \leq b$. Assume also that their dependency graph has degree $\Delta$. Then, the sum $s_N:=\sum_{i=1}^N x_i$ satisfies for every integer $M\geq \Delta+1$ and $t>0$
\begin{equation}\label{eq:hoeff_dd}
\pr\left(|s_N-\mathbb{E}[s_N]|\geq t\right)\leq 2M \exp \left( -\frac{2t^2}{N^2(b-a)^2} \left\lfloor \frac{N}{M}\right\rfloor \right).
\end{equation}
\end{corollary}

Having presented the necessary tools, the proof of Theorem
\ref{th:rip_volterra} is presented next.

\begin{IEEEproof}[Proof of Theorem \ref{th:rip_volterra}]
Consider a specific realization of $\mathbft{X}$ and its Grammian
$\mathbft{R}$. As guaranteed by the Ger\v{s}gorin disc theorem, if
$|\tilde{R}_{ii}-1|<\delta_d$ and $|\tilde{R}_{ij}|<\delta_o/s$ for
every $i,j$ with $j\neq i$ while $\delta_d+\delta_o=\delta$ for some
$\delta\in(0,1)$, then matrix $\mathbft{X}$ possesses RIP
$\delta_s\leq \delta$ \cite{HaNo10}. Thus, the probability of
$\mathbft{X}$ not satisfying RIP of value $\delta$ can be upper
bounded as
\begin{equation}\label{eq:bound1}
\pr\left(\delta_s>\delta\right)\leq \pr\left(
\bigcup_{i=1}^{M}\left\{|\tilde{R}_{ii}-1|\geq\delta_d\right\}
~\textrm{or}~
\bigcup_{i=1}^{M}\bigcup_{\substack{j=1\\ j\neq i}}^{M} \left\{|\tilde{R}_{ij}|\geq\frac{\delta_o}{s}\right\}
\right).
\end{equation}
Apparently, the events in the right-hand side (RHS) of
\eqref{eq:bound1} are not independent. Exploiting the symmetry of
$\mathbft{R}$, the union bound can be applied for only its lower
triangular part yielding
\begin{equation}\label{eq:bound2}
\pr\left(\delta_s>\delta\right)\leq
\sum_{i=2}^{M} \pr\left( |\tilde{R}_{ii}-1|\geq\delta_d\right) + \sum_{i=1}^{M}\sum_{j=i+1}^{M} \pr\left(|\tilde{R}_{ij}|\geq\frac{\delta_o}{s}
\right).
\end{equation}
Our next goal is to upper bound the probabilities appearing in the
RHS of \eqref{eq:bound2}. Different from the analysis in
\cite{HaNo10} for the linear case, the entries of $\mathbft{R}$
exhibit different statistical properties depending on the components
(constant, linear, quadratic, bilinear) of the nonlinear system they
correspond to. To signify the difference, we will adopt the notation
$\tilde{R}_{ij}^{\alpha\beta}$ instead of $\tilde{R}_{ij}$, where
$\alpha$ and $\beta$ can be any of $\{c,l,q,b\}$, to indicate that
the entry $R_{ij}^{\alpha\beta}$ is the inner product between the
$i$-th and the $j$-th columns of $\mathbft{X}$, but also the
$i$-th($j$-th) column comes from the $\alpha$($\beta$) part of the
system. For example, the element $\tilde{R}_{ij}^{ql}$ is the inner
product of a column of $\mathbft{X}^q$ with a column of
$\mathbft{X}^l$. Recall also that $\mathbft{R}$ satisfies the
requirements $\mathbb{E}[\tilde{R}_{ii}]=1$ and
$\mathbb{E}[\tilde{R}_{ij}]=0$ for $j\neq i$.

We start with the $L$ diagonal entries $\tilde{R}_{ii}^{ll}$, where
each one of them can be expressed as $\frac{3}{N}\sum_{k=1}^N
x_{n-k}^2$ for some $n$. Upon recognizing this quantity as a sum of
$N$ independent random variables confined in the interval
$\left[0,\frac{3}{N}\right]$, Hoeffding's lemma can be readily
applied. The bound obtained is multiplied by $L$ to account for all
$\tilde{R}_{ii}^{ll}$'s; hence
\begin{equation}\label{eq:bound_Riill}
\sum_{i=2}^{L+1} \pr\left( |\tilde{R}_{ii}^{ll}-1|\geq \delta_d\right) \leq 2L \exp\left( -\frac{2N\delta_d^2}{9}\right).
\end{equation}
Similarly, each one of the $L$ diagonal entries
$\tilde{R}_{ii}^{qq}$ is equal to $\frac{45}{4N}\sum_{k=1}^N
\left(x_{n-k}^2-\frac{1}{3}\right)^2$ for some $n$, which is a sum
of $N$ independent random variables bounded in
$\left[0,\frac{5}{N}\right]$. Lemma \ref{le:hoeff} yields
\begin{equation}\label{eq:bound_Riiqq}
\sum_{i=L+2}^{2L+1} \pr\left( |\tilde{R}_{ii}^{qq}-1|\geq\delta_d\right) \leq 2L \exp\left( -\frac{2N\delta_d^2}{25}\right).
\end{equation}

Before proceeding with the bilinear diagonal entries, let us
consider first the off-diagonal entries $\tilde{R}_{ij}^{ll}$. Each
one of them is a sum of the form
$\frac{3}{N}\sum_{k=1}^Nx_{n-k}x_{n-m-k}$ for $m\neq n$. However,
the summands are not generally independent; every summand is a
two-variable monomial and a single $x_n$ may appear in two summands.
This was handled in \cite{HaNo10} after proving that
$\tilde{R}_{ij}^{ll}$ can always be split into two partial sums,
each including independent terms. As a clarifying example, the entry
$\tilde{R}_{23}^{ll}$ can be expressed as
$\frac{3}{N}\left[\left(x_0x_{-1} + x_2x_1+\ldots\right) +
\left(x_1x_0 + x_3x_2+\ldots\right) \right]$. Moreover, the two
partial sums contain $\left\lfloor \frac{N}{2} \right\rfloor$ and
$\left\lceil \frac{N}{2}\right\rceil$ summands. Applying Lemma
\ref{le:hoeff_d} for $t=\delta_o/s$, $M=2$, $N_{\min}=\left\lfloor
\frac{N}{2} \right\rfloor$, and $b=-a=3/N$, it follows that
\begin{equation}\label{eq:bound_split2}
\pr\left( |\tilde{R}_{ij}^{ll}|\geq\frac{\delta_o}{s}\right) \leq 4 \exp\left(-\left\lfloor \frac{N}{2}\right\rfloor \frac{\delta_o^2}{18s^2}\right).
\end{equation}
Taking into account that $\left\lfloor \frac{N}{2}\right\rfloor \geq
\frac{N}{3}$ for $N\geq 160$, and since there are $L(L-1)/2<L^2/2$
off-diagonal $\tilde{R}_{ij}^{ll}$ terms, their collective
probability bound is
\begin{equation}\label{eq:bound_Rijll}
\sum_{i=2}^{L+1}\sum_{j=i+1}^{L+1} \pr\left( |\tilde{R}_{ij}^{ll}|\geq\frac{\delta_o}{s}\right) \leq 2L^2 \exp\left(- \frac{N\delta_o^2}{54s^2}\right).
\end{equation}

Returning to the bilinear diagonal entries, every
$\tilde{R}_{ii}^{bb}$ can be written as
$\frac{9}{N}\sum_{k=1}^Nx_{n-k}^2x_{n-m-k}^2$ for some $m\neq 0$.
Even though the summands are not independent, they exhibit identical
structural dependence observed in $\tilde{R}_{ii}^{ll}$'s; thus, the
same splitting trick can be applied here too. Upon using Lemma
\ref{le:hoeff_d} for $t=\delta_d$, $M=2$, $N_{\min}=\left\lfloor
\frac{N}{2} \right\rfloor$, $a=0$, and $b=9/N$, and adding the
contribution of all $L(L-1)/2<L^2/2$ bilinear diagonal entries, we
end up with
\begin{equation}\label{eq:bound_Riibb}
\sum_{i=2L+2}^{M}\pr\left( |\tilde{R}_{ii}^{bb}-1|\geq\delta_d\right) \leq 2L^2 \exp\left(-\frac{2N\delta_d^2}{243}\right).
\end{equation}

Regarding the entries $\tilde{R}_{1j}^{cl}$ and
$\tilde{R}_{1j}^{cq}$, an immediate application of Hoeffding's
inequality yields
\begin{equation}\label{eq:bound_Rijcl}
\sum_{j=2}^{L+1}\pr\left( |\tilde{R}_{1j}^{cl}|\geq\frac{\delta_o}{s}\right) \leq 2L \exp\left(-\frac{N\delta_o^2}{6s^2}\right)
\end{equation}
\begin{equation}\label{eq:bound_Rijcq}
\sum_{j=L+2}^{2L+1}\pr\left( |\tilde{R}_{1j}^{cq}|\geq\frac{\delta_o}{s}\right) \leq 2L \exp\left(-\frac{8N\delta_o^2}{45s^2}\right)
\end{equation}
whereas the probabilities $\pr\left(|\tilde{R}_{1j}^{cb}|\geq
\delta_o/s\right)$ have been already accounted for in the analysis
of the $\tilde{R}_{ij}^{ll}$'s.

The entries $\tilde{R}_{ij}^{lq}$ can be written as
$\frac{3\sqrt{15}}{2N}\sum_{k=1}^N x_{n-k}
\left(x_{n-k-m}^2-\frac{1}{3}\right)$ for some $n$ and $m$, where
every summand lies in $\left[-\frac{\sqrt{15}}{N},
\frac{\sqrt{15}}{N}\right]$. Two sub-cases will be considered. The
first corresponds to the $L$ entries $\tilde{R}_{ij}^{lq}$ with
$m=0$ (or equivalently $j=i+L$), in which every summand depends on a
single input. Through Lemma \ref{le:hoeff}, the sum of probabilities
related to these $L$ entries is upper bounded by
$2L\exp(-N\delta_o^2/(30s^2))$. The second case includes the
remaining $(L^2-L)$ entries with $m\neq 0$, for which the splitting
trick can be applied to yield the bound $4(L^2-L)\exp\left(-\lfloor
N/2 \rfloor\delta_o^2/(30s^2) \right)$. Combining the two bounds
yields
\begin{equation}\label{eq:bound_Rijlq}
\sum_{i=2}^{L+1}\sum_{j=L+2}^{2L+1}\pr\left( |\tilde{R}_{ij}^{lq}|\geq\frac{\delta_o}{s}\right) \leq 4L^2 \exp\left(-\frac{N\delta_o^2}{90s^2} \right).
\end{equation}

The $\tilde{R}_{ij}^{qq}$ entries can be expressed as
$\frac{45}{4N}\sum_{k=1}^N
(x_{n-k}^2-\frac{1}{3})(x_{n-k-m}^2-\frac{1}{3})$ for some $m\neq
0$, where each summand is bounded in
$\left[-\frac{5}{2N},\frac{10}{2N}\right]$. Exploiting the same
splitting trick and summing up the contributions of all the
$L(L-1)/2$ $\tilde{R}_{ij}^{qq}$ entries, yields
\begin{equation}\label{eq:bound_Rijqq}
\sum_{i=L+2}^{2L+1}\sum_{j=i+1}^{2L+1}\pr\left( |\tilde{R}_{ij}^{qq}|\geq\frac{\delta_o}{s}\right) \leq 2L^2 \exp\left(-\frac{8N\delta_o^2}{675s^2} \right).
\end{equation}

The $\tilde{R}_{ij}^{lb}$'s can be written as the sum
$\frac{3\sqrt{3}}{N}\sum_{k=1}^N x_{n-k}x_{n-k-m}x_{n-k-p}$ for some
$n$ and  $m\neq p$, while every summand lies in
$\left[-\frac{3\sqrt{3}}{N},\frac{3\sqrt{3}}{N}\right]$. Note that
there exist $\tilde{R}_{ij}^{lb}$'s with summands being two-input
monomials, i.e., for $m=0$ or $p=0$. However, to simplify the
presentation, the derived bound is slightly loosened by considering
all $\tilde{R}_{ij}^{bl}$'s as sums of three-input monomials. This
specific structure precludes the application of the splitting
procedure into two halves, and necessitates use of the dependency
graph. It can be shown that the degree of the dependency graph
associated with the three-variable products for any
$\tilde{R}_{ij}^{lb}$ entry is at most 6. Then, application of
Corollary \ref{co:hoeff_dd} over the $L^2(L-1)/2\leq L^3/2$
$\tilde{R}_{ij}^{lb}$ entries together with the inequality
$\left\lfloor N/7 \right\rfloor\geq N/8$, which holds for $N\geq
160$, yield
\begin{equation}\label{eq:bound_Rijlb}
\sum_{i=2}^{L}\sum_{j=2L+2}^{M}\pr\left( |\tilde{R}_{ij}^{lb}|\geq\frac{\delta_o}{s}\right) \leq 7L^3 \exp\left(-\frac{N\delta_o^2}{432s^2} \right).
\end{equation}

The $\tilde{R}_{ij}^{qb}$'s can be written as
$\frac{9\sqrt{5}}{2N}\sum_{k=1}^N
\left(x_{n-k}^2-\frac{1}{3}\right)x_{n-k-m}x_{n-k-p}$ for some $n$
and $m\neq p$, where the summands lie in
$\left[-\frac{3\sqrt{5}}{N},\frac{3\sqrt{5}}{N}\right]$. Following a
reasoning similar to the one for $\tilde{R}_{ij}^{lb}$,
\begin{equation}\label{eq:bound_Rijqb}
\sum_{i=L+2}^{2L+1}\sum_{j=2L+2}^{M}\pr\left( |\tilde{R}_{ij}^{qb}|\geq\frac{\delta_o}{s}\right) \leq 7L^3 \exp\left(-\frac{N\delta_o^2}{720s^2} \right).
\end{equation}

Finally, the $\tilde{R}_{ij}^{bb}$'s are expressed as
$\frac{9}{N}\sum_{k=1}^N x_{n-k}x_{n-k-m}x_{n-k-p}x_{n-k-m-q}$ for
some $n$, $m$, $p$, and $q$, whereas the summands lie in
$\left[-\frac{9}{N},\frac{9}{N}\right]$. For any
$\tilde{R}_{ij}^{bb}$ entry, the summands are four-input monomials,
and thus, the degree of the associated dependency graph is at most
12. Upon applying Corollary \ref{co:hoeff_dd} over the
$L(L-1)(L^2-L-2)/8$ $\tilde{R}_{ij}^{bb}$'s, and since $\left\lfloor
N/13 \right\rfloor \geq N/14$ for $N\geq 160$, we obtain
\begin{equation}\label{eq:bound_Rijbb}
\sum_{i=2L+2}^{M}\sum_{j=i+1}^{M}\pr\left( |\tilde{R}_{ij}^{bb}|\geq\frac{\delta_o}{s}\right) \leq \frac{13}{4}L^4 \exp\left(-\frac{N\delta_o^2}{2268s^2} \right).
\end{equation}

Adding together the bounds for the diagonal elements
\eqref{eq:bound_Riill}, \eqref{eq:bound_Riiqq}, and
\eqref{eq:bound_Riibb}, implies
\begin{equation}\label{eq:bound_diagonal}
\sum_{i=2}^{M}\pr\left( |\tilde{R}_{ii}-1|\geq \delta_d\right) \leq 3L^2 \exp\left(-\frac{2N\delta_d^2}{243}\right)
\end{equation}
for $L\geq 7$. For the off-diagonal elements, upon adding
\eqref{eq:bound_Rijll},
\eqref{eq:bound_Rijcl}-\eqref{eq:bound_Rijbb}, it follows for $L\geq
7$ that
\begin{equation}\label{eq:bound_offdiagonal}
\sum_{i=2}^{M}\sum_{j=i+1}^{M}\pr\left( |\tilde{R}_{ij}|\geq\frac{\delta_o}{s}\right) \leq 6L^4\exp\left(-\frac{N\delta_o^2}{2268s^2} \right).
\end{equation}
By choosing $\delta_d=\frac{\delta_o}{s}\sqrt{\frac{3}{56}}$, the
arguments of the exponentials in \eqref{eq:bound_diagonal} and
\eqref{eq:bound_offdiagonal} become equal, and after adding the two
bounds, we arrive at
\begin{equation}\label{eq:bound3}
\pr\left(\delta_s>\delta\right)\leq 7L^4\exp\left(-\frac{N\delta_o^2}{2268s^2} \right).
\end{equation}
Since $\delta=\delta_d + \delta_o$ translates to $\delta_o^2=\left(
\frac{s\sqrt{56/3}}{s\sqrt{56/3}+1} \right)^2 \delta^2>0.8\delta^2$
for $s\geq 2$, the bound in \eqref{eq:bound3} simplifies to
\begin{equation}\label{eq:bound4}
\pr\left(\delta_s>\delta\right)\leq 7L^4\exp\left(-\frac{N\delta^2}{2835s^2}\right)\leq
\exp\left(-\frac{N\delta^2}{s^2}\left(\frac{1}{2835}-\frac{5s^2}{N\delta^2}\log L
\right)\right).
\end{equation}
Now set $C{:=}2,835$ and choose any $\gamma\in(0,1)$. Whenever
$N\geq \frac{5C}{(1-\gamma)\delta^2}\cdot s^2 \log L$,
\eqref{eq:bound4} yields
\begin{equation*}
\pr\left(\delta_s>\delta\right)\leq
\exp\left(- \frac{\gamma\delta^2}{C}\cdot \frac{N}{s^2} \right)
\end{equation*}
which completes the proof.
\end{IEEEproof}

\section*{Acknowledgments}
The authors would like to thank Dr. Daniele Angelosante and Prof.
Xiaodonog Cai for valuable feedback on the contributions of this
paper.

\bibliographystyle{IEEEtranS}
\bibliography{IEEEabrv,volterra}

\begin{figure*}[h]
\begin{minipage}[t]{3in}
\begin{algorithm}[H]
\caption{CCD-(W)L}
\label{alg:batch}
\begin{algorithmic}[1]
\linespread{0.9}
\footnotesize
\renewcommand{\arraystretch}{0.5}
\STATE Initialize $\mathbf{z}=\mathbf{X}^T\mathbf{y}$.
\STATE Compute matrix $\mathbf{R}=\mathbf{X}^T\mathbf{X}$.
\REPEAT
\FOR{$i=1,\ldots,M$}
\STATE Update $\mathbf{z}$ as $\mathbf{z} = \mathbf{z} + \mathbf{r}_i \hat{h}_i$.
\STATE Update $\hat{h}_i$ using \eqref{eq:hi_formula}.
\STATE Update $\mathbf{z}$ as $\mathbf{z} = \mathbf{z} - \mathbf{r}_i \hat{h}_i$.
\ENDFOR
\UNTIL{convergence of $\hat{\mathbf{h}}$.}
\end{algorithmic}
\end{algorithm}
\end{minipage}
\hfill
\begin{minipage}[t]{3in}
\begin{algorithm}[H]
\caption{CCD-R(W)L}
\label{alg:recursive}
\begin{algorithmic}[1]
\linespread{0.9}
\footnotesize
\renewcommand{\arraystretch}{0.5}
\STATE Initialize $\hat{\mathbf{h}}_0=\mathbf{0}_{M}$, $\mathbf{z}_0=\mathbf{0}_{M}$,
$\mathbf{R}_0=\delta \mathbf{I}_M$.
\FOR{$N=1,2,\ldots$}
\STATE Update $\mathbf{R}_N$ and $\mathbf{z}_N$ via \eqref{eq:R_update} and \eqref{eq:z_update}.
\FOR{$i=1,\ldots,M$}
\STATE $\mathbf{z}_N = \mathbf{z}_N + \mathbf{r}_{N,i} \hat{h}_{N-1,i}$
\STATE $\hat{h}_{N,i}= \frac{\sign(z_{N,i})}{R_{N,ii}}\cdot \left[|z_{N,i}|-\lambda_N w_{N,i}\right]_+$
\STATE $\mathbf{z}_N=\mathbf{z}_{N} - \mathbf{r}_{N,i} \hat{h}_{N,i}$
\ENDFOR
\ENDFOR
\end{algorithmic}
\end{algorithm}
\end{minipage}
\hfill
\end{figure*}


\begin{table}
\caption{Experimental results for synthetic and real QTL data}
\label{tbl:QTLtables}
\centering
\subtable[Synthetic data]{
\centering
\renewcommand{\arraystretch}{0.8}
\begin{tabular}{|l|r|r|r|r|}
\hline
\textbf{Method} & \textbf{PE} & \textbf{MSE} & \textbf{NNZ} & $\delta$/$\lambda$\\
\hline
Ridge           & 68.10 & 82.29 & 7382 & 0.61 $ N$\\
Lasso       &   12.84   &   15.85   &   200 &   0.19 $N$\\
wLasso  &   13.09   &   5.11    &   85  &   3.77 $N$\\
\hline
\end{tabular}\label{tbl:synthetic}}
\subtable[Real QTL barley data]{
\centering
\renewcommand{\arraystretch}{0.8}
\begin{tabular}{|l|r|r|r|}
\hline
\textbf{Method} & \textbf{PE} & \textbf{NNZ} & $\delta$/$\lambda$\\
\hline
Ridge           & 8.26  & 8129 & 4.28$\cdot 10^{4}$ $ N$\\
Lasso       &   5.96    &   48  &   0.33 $N$\\
wLasso  &   5.69    &   34  &   6.88 $N$\\
\hline
\end{tabular}\label{tbl:real}}
\centering
\subtable[QTLs estimated by wLasso for the real barley data]{
\centering
\renewcommand{\arraystretch}{0.8}
\begin{tabular}{|lr|lr|}
\hline
\multicolumn{2}{|c|}{\textbf{Main effects}} & \multicolumn{2}{|c|}{\textbf{Epistatic effects}} \\
\hline
Marker & Value &Markers & Value\\
\hline
(12)    & $+0.78$   & (7,66)    & $+0.19$   \\
(53)    & $-0.18$   & (9,33)    & $-0.29$   \\
(61)    & $+0.23$   & (20,95)   & $+0.13$   \\
(101)   & $+0.40$   & (33,88)   & $+0.10$   \\
(104)   & $+0.24$   & (35,99)   & $-0.47$   \\
(112)   & $+0.43$   & (38,52)   & $-0.15$   \\
            &                   & (56,92)   & $+0.38$   \\
            &                   & (63,81)   & $-0.19$   \\
\hline
\end{tabular}\label{tbl:real2}}
\end{table}


\end{document}